\pdfoutput=1
\documentclass[10pt]{nus-paper}

\usetikzlibrary{arrows.meta,positioning,calc}
\usepackage{enumitem}
\usepackage{etoolbox}
\usepackage{relsize}
\AtBeginEnvironment{tabular}{\fontfamily{ptm}\selectfont}
\AtBeginEnvironment{tabularx}{\fontfamily{ptm}\selectfont}
\AtBeginEnvironment{tabular*}{\fontfamily{ptm}\selectfont}
\DeclareCaptionFont{tabtimes}{\fontfamily{ptm}\selectfont}
\definecolor{omniO}{HTML}{EF476F}
\definecolor{omniM}{HTML}{F28E2B}
\definecolor{omniN}{HTML}{00A890}
\definecolor{omniI}{HTML}{934BED}
\definecolor{omniSci}{HTML}{184878}

\usepackage{xspace}
\newcommand{\OmniName}{\textcolor{omniO}{O}\textcolor{omniM}{m}\textcolor{omniN}{n}\textcolor{omniI}{i}\textcolor{omniSci}{Scientist}\xspace}
\usepackage{fontawesome5}
\newcommand{\evPerc}{\textcolor{omniN}{\faEye}}
\newcommand{\evSym}{\textcolor{omniI}{\faCode}}
\newcommand{\evQuant}{\textcolor{omniM}{\faChartBar}}
\newcommand{\evProc}{\textcolor{omniO}{\faStream}}
\newcommand{\mdmark}[1]{\textcolor{nusblue}{#1}}
\newcommand{\mdImage}{\mdmark{\faImage}}
\newcommand{\mdSignal}{\mdmark{\faWaveSquare}}
\newcommand{\mdSpectrum}{\mdmark{\faChartArea}}
\newcommand{\mdAudio}{\mdmark{\faVolumeUp}}
\newcommand{\mdVideo}{\mdmark{\faVideo}}
\newcommand{\mdThreeD}{\mdmark{\faCube}}
\newcommand{\mdTraj}{\mdmark{\faRoute}}
\newcommand{\mdTable}{\mdmark{\faTable}}
\newcommand{\mdFormula}{\mdmark{\faSuperscript}}
\newcommand{\mdSeq}{\mdmark{\faDna}}
\newcommand{\mdField}{\mdmark{\faBorderAll}}
\newcommand{\mdGraph}{\mdmark{\faProjectDiagram}}

\extrafloats{100}

\definecolor{tblZebra}{HTML}{EAECEF}
\definecolor{tblHi}{HTML}{E7F4F0}
\definecolor{tblBlue}{HTML}{1A73E8}
\definecolor{tblGain}{HTML}{1B7F5C}
\newcolumntype{Y}{>{\centering\arraybackslash}X}
\AtBeginDocument{\setlength{\aboverulesep}{0pt}\setlength{\belowrulesep}{0pt}\setlength{\extrarowheight}{1.5pt}}
\newcommand{\best}[1]{{\bfseries\boldmath\textcolor{tblGain}{\textscale{1.09}{#1}}}}

\newcommand{\hib}{{\scriptsize$\uparrow$}}
\newcommand{\lob}{{\scriptsize$\downarrow$}}
\newcommand{\stkhd}[2]{\begin{tabular}{@{}c@{}}\textbf{#1}\\\textbf{#2}\end{tabular}}

\definecolor{prmBg}{HTML}{EAF1FA}
\definecolor{prmBar}{HTML}{D3E2F4}
\definecolor{prmRule}{HTML}{A6BFDD}
\definecolor{prmInk}{HTML}{1B3A5C}
\newcommand{\prmtitle}[1]{%
  \par\addvspace{9pt}\noindent
  \colorbox{prmBar}{\makebox[\dimexpr\linewidth-2\fboxsep\relax][l]{%
    \rmfamily\bfseries\small\textcolor{prmInk}{\strut #1}}}%
  \par\nobreak\vspace{-\parskip}}
\lstdefinestyle{prompt}{%
  basicstyle={\rmfamily\small\linespread{1.08}\selectfont},
  numbers=none,
  xleftmargin=8.8pt, framexleftmargin=8.8pt,
  xrightmargin=6pt, framexrightmargin=6pt,
  frame=none, framesep=6pt, framextopmargin=7pt, framexbottommargin=7pt,
  backgroundcolor=\color{prmBg},
  columns=fullflexible, keepspaces=true,
  showstringspaces=false, breaklines=true, breakindent=0pt, tabsize=2,
  captionpos=b, aboveskip=0pt, belowskip=3pt,
  literate={-}{{\mbox{-}}}1 {"}{{\textquotedbl}}1 {'}{{\textquotesingle}}1,
  moredelim=**[is][\bfseries]{@}{@},
  moredelim=**[is][\color{nuswarm}]{$}{$}}
\newcommand{\zb}{\rowcolor{tblZebra}}
\newcommand{\grpfour}[2]{\addlinespace[2pt]\midrule\rowcolor{tblZebra}\multicolumn{4}{@{}l}{\textbf{#1}} & \textcolor{nusgray}{#2}\\}
\newcommand{\grpfourA}[2]{\rowcolor{tblZebra}\multicolumn{4}{@{}l}{\textbf{#1}} & \textcolor{nusgray}{#2}\\}
\newcommand{\grpspan}[2]{\addlinespace[2pt]\midrule\rowcolor{tblZebra}\multicolumn{#1}{@{}l}{\textbf{#2}}\\}
\newcommand{\grpspanA}[2]{\rowcolor{tblZebra}\multicolumn{#1}{@{}l}{\textbf{#2}}\\}

\begin{document}
\arrayrulecolor{black}%

\setrunninghead{OmniScientist: An Omni-Modal Omni-Discipline AI Scientist}

\setrunningauthor{Li et al.}

\title{\OmniName:\\ An Omni-Modal Omni-Discipline AI Scientist}

\author{Bobo Li\textsuperscript{1}} \author{Hao Fei\textsuperscript{2*}} \author{Tianjie Ju\textsuperscript{1}} \author{Mong-Li Lee\textsuperscript{1}} \author{Wynne Hsu\textsuperscript{1}} \affil{\textsuperscript{1}National University of Singapore \quad \textsuperscript{2}University of Oxford} \affil{\normalfont\upshape\fontsize{10.5}{12.5}\selectfont Project page: \url{https://omni-scientist.github.io}} \affil{\normalfont\upshape\fontsize{10.5}{12.5}\selectfont Software \& Skill: \url{https://github.com/Omni-Scientist/OmniScientist}}

\firstpagenote{Bobo Li (\texttt{libobo@nus.edu.sg}), Hao Fei (Correspondence/Project Lead, \texttt{haofei7419@gmail.com}).}

\maketitle

\begin{abstract}
Recent advances in foundation models have enabled AI scientists to automate increasingly complete research workflows, from hypothesis generation and code execution to manuscript preparation. Yet workflow coverage alone does not provide access to the full evidence on which scientific discovery depends. Existing systems typically reason over text, code, labels, or precomputed summaries, leaving scientifically decisive spatial, temporal, cross-channel, and procedural relations unavailable to the agent. We introduce \OmniName, an end-to-end, omni-modal AI scientist that conducts multidisciplinary research directly from heterogeneous raw evidence. A perception layer and 3 autonomous agents for ideation, experiment, and writeup operate within a deterministic pipeline, allowing observations to shape research questions, experimental decisions, and final claims throughout the research lifecycle. By running idea, rigour, and claim checks in code, the system enforces novelty screening, statistical validity, execution provenance, and numerical traceability. We evaluate \OmniName on 36 real-data cases spanning 5 discipline families, 4 families of scientific evidence, and modalities including images, signals, audio, video, 3-D structures, trajectories, tables, formulae, and graphs. The system completes the full path from raw data to a compiled manuscript in all 36 cases and achieves a mean overall paper score of $6.3$ with the reference reasoning backbone. In paired comparisons against a blind variant that receives only precomputed scalar features, direct perception improves all 7 evaluation dimensions and wins $85\%$ of head-to-head judgments. These results show that lifecycle-wide perception is essential to the evidence-grounded scientific discovery and provides a practical path toward broadly capable AI scientists.
\end{abstract}


{\setlength{\parskip}{0pt}\setlength{\abovecaptionskip}{7pt}\centering
  \includegraphics[width=0.98\linewidth]{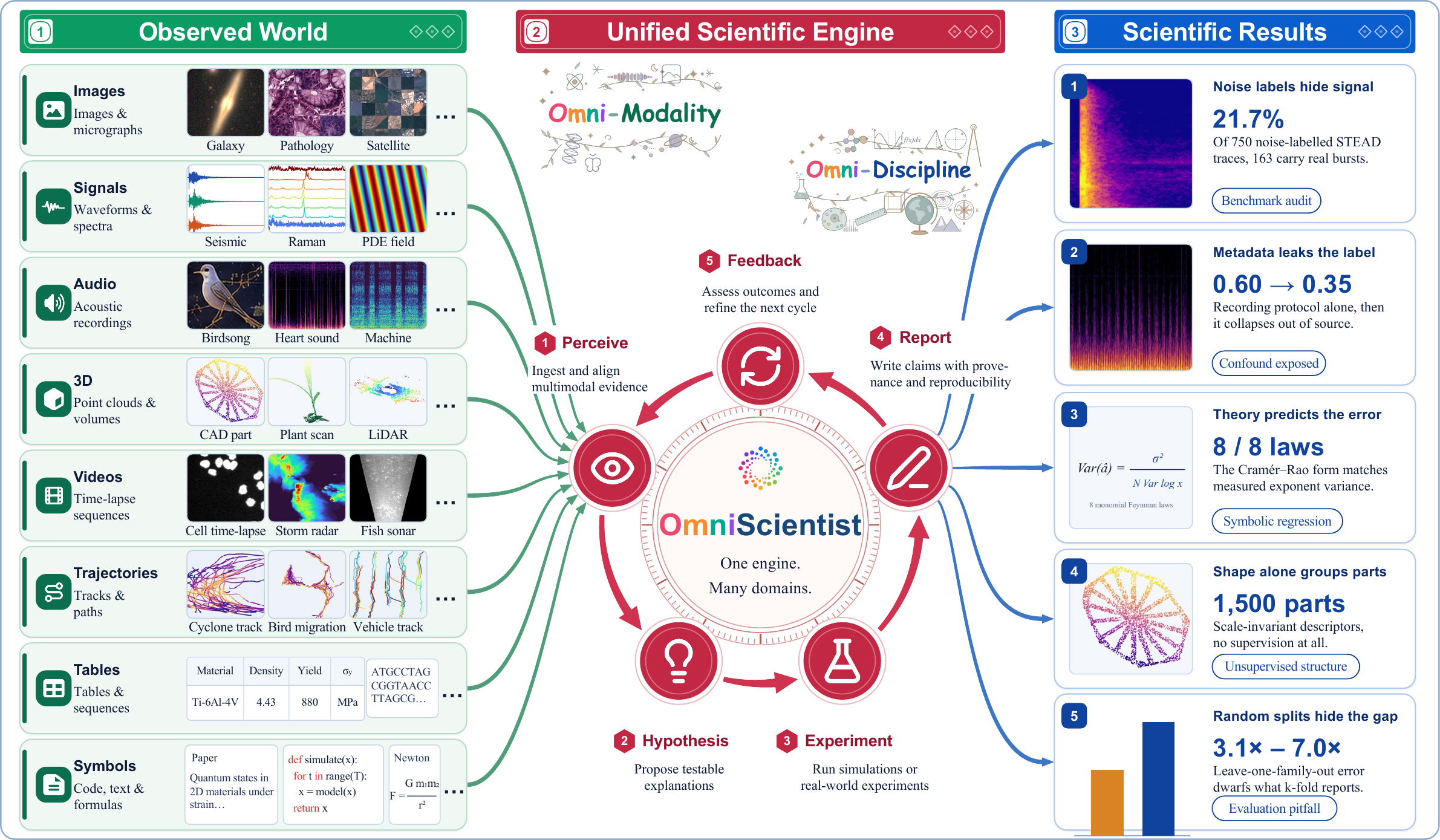}
  \captionof{figure}{Overview of the \OmniName framework. Raw multimodal observations spanning multiple disciplines (left) are fed into a unified engine that perceives evidence, executes falsifiable experiments, and grounds every claim (center). This continuous pipeline ultimately yields verifiable scientific discoveries across various fields (right).}
  \label{fig:teaser}\par}

\section{Introduction}
\label{sec:intro}

Recent advances in foundation models have expanded the scope of automated scientific discovery.
Improvements in reasoning, code generation, and tool use \citep{openai2023gpt4, anthropic2024claude3, deepseekr1} have moved scientific AI from task-specific systems, exemplified by protein structure prediction \citep{jumper2021alphafold}, toward agents that coordinate hypothesis formulation, experiment execution, and scientific writing \citep{schmidgall2025agentlab, yuan2025dolphin, tang2025airesearcher}.
At the current frontier, end-to-end systems can start from a research direction and autonomously produce candidate ideas, runnable code, experimental figures, complete manuscripts, and self-reviews \citep{lu2026aiscientist, yamada2025aiscientist2, internagent2025, intology2025zochi, gottweis2026coscientist}.
These systems therefore cover almost the entire visible research workflow.
A separate capability remains unresolved: an AI scientist must also derive defensible questions and claims from the heterogeneous observations on which science depends.

Scientific evidence reaches researchers in forms with markedly different internal structure.
Text, formulae, sequences, and knowledge graphs encode symbolic relations, while microscopy, spectra, waveforms, audio, video, 3-D structures, distributions, and trajectories carry spatial, temporal, cross-channel, statistical, and procedural relations \citep{wang2023scidiscovery, baltrusaitis2019multimodal, yue2023mmmu, li2024mmsci}.
All of these artifacts can be serialised into tokens or accessed through code.
The decisive issue is which relations survive the interface.
A caption can omit local morphology, an unordered feature vector can erase temporal order, and a small set of scalars can hide cross-channel inconsistency.
Yet most current AI-scientist pipelines expose data through text, code, labels, or summaries prepared in advance \citep{lu2026aiscientist, yamada2025aiscientist2, jiang2025aide, schmidgall2025agentlab, yuan2025dolphin, internagent2025}.
The agent therefore inherits a human-chosen representation before inquiry begins.
This interface constrains the anomalies it can notice, the hypotheses it can formulate, and the claims it can support.
Existing systems are increasingly \emph{workflow-complete}, while remaining \emph{evidence-incomplete}.

Strong multimodal models do not by themselves close this gap.
Scientific multimodal benchmarks usually fix the observation and the question in advance \citep{wang2024charxiv, roberts2024scifibench, pramanick2024spiqa, li2024mmsci}, while scientific agents that use perception generally invoke it at a local stage, such as inspecting generated figures, scoring plots, reading an interface, or monitoring an apparatus \citep{yamada2025aiscientist2, gandhi2025cmbagentvlm, sun2025scienceboard, darvish2024organa, mandal2025aila}.
Such uses can improve an isolated decision, yet they do not establish a research loop in which an observation can redirect the study.
In a perception-driven AI scientist, raw evidence should influence which question is selected, how the experiment is designed, what is examined after execution, and which statements are admitted to the final manuscript.
A broader observation and search space also enlarges the room for data leakage, repeated testing, hypothesising after results are known (HARKing), and unsupported reporting.
Lifecycle-wide perception therefore needs a control structure that preserves provenance and enforces statistical and factual constraints.

To meet these requirements, we introduce \OmniName (cf. \cref{fig:teaser}), an end-to-end, omni-modal AI scientist for multidisciplinary discovery.
A task is specified by a dataset, a scientific subject, and a target property, together with the raw artifacts referenced by that specification.
\OmniName coordinates a perception layer with 3 autonomous agents for ideation, experiment, and writeup (\cref{fig:arch}).
Within each stage, a ReAct loop interleaves observation, reasoning, and action \citep{yao2023react, shinn2023reflexion, schick2023toolformer}, allowing evidence to shape question formation, experimental design, result inspection, and scientific argumentation.
Around these open-ended agents, a thin deterministic pipeline controls stage transitions, returns to ideation when an experiment collapses or yields a null result, and admits each stage output only after a check enforced in code.
The idea, rigour, and claim checks examine novelty evidence and falsifiability, leakage and effective sample size, execution provenance and multiple comparisons, anti-HARKing constraints, numerical traceability, and claim support.
We further organise scientific evidence into 4 discipline-independent families, namely perceptual, symbolic, quantitative-statistical, and procedural evidence (cf. \cref{tab:evidence}).
The same engine can consequently operate on a seismogram, a CAD mesh, or a knowledge graph.
Adding a discipline requires a new specification file, with no change to the core pipeline and no domain-specific research code.

We evaluate \OmniName on a 36-case demonstration suite spanning 5 discipline families, all 4 evidence families, and modalities ranging from images and waveforms to audio, video, 3-D structures, trajectories, tables, formulae, and graphs (\cref{tab:disciplines}).
The system completes the full path from raw data to a compiled paper in all 36 cases.
With the reference reasoning backbone, the generated papers receive a mean overall score of $6.3$ on a 7-dimensional rubric scored by 2 cross-family judges, and paper quality remains broadly consistent across evidence modalities (\cref{tab:eval-backbone,tab:eval-aggregate}).
We then isolate the contribution of perception by comparing the full system with a blind variant that receives precomputed scalar features and never accesses the raw observation.
Full perception improves every evaluated dimension, with the largest gains in multimodal grounding and scientific significance, and its papers win $85\%$ of the head-to-head judgments (\cref{fig:eval-gain}).
The gains appear in the scientific substance of the papers: the questions selected, the analyses executed, and the claims supported by the resulting evidence.

In summary, our main contributions are as follows:
\begin{itemize}[leftmargin=1.2em, itemsep=3pt, topsep=3pt]
\item We present \OmniName, an end-to-end, omni-modal, and discipline-agnostic AI scientist that works directly from heterogeneous scientific evidence. It extends automated discovery beyond workflow coverage by keeping spatial, temporal, statistical, and procedural relations available throughout the research process.

\item We propose a perception-first, multi-agent architecture where raw observations actively guide ideation, experimentation, and manuscript generation. Furthermore, we integrate code-enforced checks across the pipeline to screen for novelty, ensure statistical validity and execution provenance, prevent HARKing, and guarantee that all reported claims are traceable and supported by evidence.

\item We establish a 36-case demonstration suite across 5 discipline families and 4 families of scientific evidence. The 36 completed end-to-end runs, together with backbone comparisons and a paired blind ablation, demonstrate broad cross-disciplinary applicability and show that direct perception is essential to the evidence-grounded scientific discovery.
\end{itemize}

\section{Related work}
\label{sec:related}
\subsection{Autonomous agents for scientific discovery}
\label{sec:rel-agents}
Autonomous agents are increasingly participating in scientific discovery.
Early frameworks targeted one domain and one instrument, executing laboratory procedures or choosing the next synthesis from diffraction patterns \citep{boiko2023coscientist, szymanski2023alab}.
Later systems widened to real scientific data across disciplines, proposing hypotheses that were subsequently validated at the bench \citep{mitchener2025kosmos, villaescusa2025denario, swanson2025virtuallab, gottweis2026coscientist, ghareeb2026robin}.
A parallel wave automated the entire research lifecycle on software repositories, from idea generation to a drafted and self-reviewed manuscript, with an improved numerical metric as the measure of success \citep{lu2026aiscientist, yamada2025aiscientist2, jiang2025aide, schmidgall2025agentlab, yuan2025dolphin, internagent2025}, most recently embedding that pipeline in a collaborative ecosystem \citep{shao2025omniscientist}.
Despite these advances, current agents attend only to what a person has already turned into text, code, or a number, and overlook the raw evidence itself.
As a result, almost every benchmark grades the artifacts a system produces and never what it can perceive \citep{chan2024mlebench, starace2025paperbench, chen2024scienceagentbench, wei2025agenticscience}.
We therefore build an AI scientist that perceives the raw observation directly, across the 4 families of scientific evidence in \cref{tab:evidence}.

\subsection{Multimodal perception for science}
\label{sec:rel-perception}
General multimodal models already provide strong perceptual capability, through contrastive image-text pretraining \citep{radford2021clip} and instruction-following vision-language models \citep{alayrac2022flamingo, liu2023llava}.
Scientific benchmarks have measured that capability on published charts and figures \citep{wang2024charxiv, roberts2024scifibench, pramanick2024spiqa, li2024mmsci}, on microscopy and materials observations \citep{lozano2024mubench, burgess2025microvqa, alampara2025macbench, zhou2025sfe}, and on experiment video and bioacoustics \citep{xu2025expvid, robinson2024naturelm}.
Yet they treat perception as a standalone question-answering skill, with a human selecting the observation, posing the question, and judging the response.
Even MicroVQA scores hypothesis generation and experiment proposal against fixed choices, and instrument-specific and pathology models remain predictors for tasks defined in advance \citep{chen2024uni}.
Scientific agents apply perception more directly, though only at isolated stages, inspecting their own figures \citep{yamada2025aiscientist2}, scoring generated plots \citep{gandhi2025cmbagentvlm}, reading a software interface \citep{sun2025scienceboard}, watching an apparatus mid-experiment \citep{darvish2024organa}, or driving a microscope from fixed thresholds \citep{mandal2025aila}.
A few systems let observations inform claim formation \citep{yao2025scilink, zhao2026discoper}, though perception still plays a local and fragmented role in the research process.
\OmniName makes perception active at every stage, from forming the question to grounding the claims in the finished paper.

\begin{table}[p]
  \centering
  \caption{
  The demonstration suite: 5 categories, 36 cases, one real downloadable dataset each, with the sample count $N$ per dataset.
  \emph{Evidence}~--~\evPerc\,perceptual, \evSym\,symbolic, \evQuant\,quantitative-statistical, \evProc\,procedural.
  \emph{Modality}~--~\mdImage\,image, \mdSignal\,signal, \mdSpectrum\,spectrum, \mdAudio\,audio, \mdVideo\,video,
  \mdThreeD\,3-D, \mdTraj\,trajectory, \mdTable\,table, \mdFormula\,formula, \mdSeq\,sequence, \mdField\,field, \mdGraph\,graph.
    }
  \label{tab:disciplines}
  \renewcommand{\arraystretch}{1.08}\setlength{\tabcolsep}{6pt}
  \begin{tabularx}{\linewidth}{@{}>{\raggedright\arraybackslash}p{4.05cm}X >{\centering\arraybackslash}p{1.5cm} >{\centering\arraybackslash}p{1.5cm} r@{}}
    \toprule
    \textbf{Discipline} & \textbf{Representative dataset} & \textbf{Evidence} & \textbf{Modality} & \textbf{$N$} \\
    \midrule
    \grpfourA{Physical sciences}{5 cases}
    Condensed matter / nano & NFFA-EUROPE~\citep{data_nffa_sem} & \evPerc & \mdImage & 2{,}655 \\
    \zb Vibrational spectroscopy & RRUFF~\citep{data_rruff} & \evPerc & \mdSpectrum & 2{,}000 \\
    Materials informatics & UCI superconductor~\citep{data_supercon} & \evQuant & \mdTable & 21{,}263 \\
    \zb Molecular chemistry & PubChem~\citep{data_pubchem} & \evPerc & \mdImage & 30 \\
    Symbolic regression & Feynman~\citep{data_feynman} & \evSym & \mdFormula & 12 \\
    \grpfour{Earth \& space}{9 cases}
    Remote sensing & EuroSAT~\citep{data_eurosat} & \evPerc & \mdImage & 5{,}000 \\
    \zb Galaxy morphology & Galaxy Zoo~\citep{data_galaxyzoo} & \evPerc & \mdImage & 1{,}000 \\
    Galaxy cross-survey & GZ DECaLS~\citep{data_gzdecals} & \evPerc & \mdImage & 210 \\
    \zb Gravitational waves & GWOSC~\citep{data_gwosc} & \evPerc & \mdSignal & 1{,}500 \\
    Seismology & STEAD~\citep{data_stead} & \evPerc & \mdSignal & 1{,}500 \\
    \zb Marine biology & WHOI-Plankton~\citep{data_whoiplankton} & \evPerc & \mdImage & 3{,}000 \\
    Geology / petrophysics & Digital Rocks~\citep{data_digitalrocks} & \evPerc & \mdThreeD & 375 \\
    \zb Meteorology & SEVIR~\citep{data_sevir} & \evPerc & \mdVideo & 384 \\
    Cyclone dynamics & IBTrACS~\citep{data_ibtracs} & \evProc & \mdTraj & 400 \\
    \grpfour{Life \& medical}{7 cases}
    Pathology & Kather CRC~\citep{data_kather} & \evPerc & \mdImage & 5{,}000 \\
    \zb Radiology & Chest X-ray~\citep{data_kermany} & \evPerc & \mdImage & 3{,}000 \\
    Medical imaging & MedMNIST CT~\citep{data_medmnist} & \evPerc & \mdThreeD & 1{,}496 \\
    \zb Cardiology & CinC 2016~\citep{data_heartsound} & \evPerc & \mdAudio & 2{,}000 \\
    Sleep neuroscience & Sleep-EDF~\citep{data_sleepedf} & \evPerc & \mdSignal & 1{,}520 \\
    \zb Cell biology & Cell Tracking Ch.~\citep{data_celltrack} & \evPerc & \mdVideo & 280 \\
    Genomics & DNA (H3)~\citep{data_dna_histone} & \evSym & \mdSeq & 10{,}000 \\
    \grpfour{Agricultural \& ecological}{8 cases}
    Plant pathology & PlantVillage~\citep{data_plantvillage} & \evPerc & \mdImage & 3{,}002 \\
    \zb Precision agriculture & Indian Pines~\citep{data_indianpines} & \evPerc & \mdSpectrum & 2{,}000 \\
    Animal behavior & CalMS21~\citep{data_calms21} & \evPerc & \mdImage & 2{,}500 \\
    \zb Ecoacoustics & Bird Audio Det.~\citep{data_birdaudio} & \evPerc & \mdAudio & 2{,}000 \\
    Marine bioacoustics & Watkins MMSD~\citep{data_watkins} & \evPerc & \mdAudio & 1{,}697 \\
    \zb Plant phenotyping & Pheno4D~\citep{data_pheno4d} & \evPerc & \mdThreeD & 223 \\
    Fisheries & Caltech Fish~\citep{data_cfc} & \evPerc & \mdVideo & 120 \\
    \zb Movement ecology & White-stork GPS~\citep{data_whitestork} & \evProc & \mdTraj & 49 \\
    \grpfour{Engineering \& information}{7 cases}
    Mechanical / mfg. & MCB~\citep{data_mcb} & \evPerc & \mdThreeD & 1{,}500 \\
    \zb Civil / surveying & SemanticKITTI~\citep{data_semantickitti} & \evPerc & \mdThreeD & 1{,}200 \\
    Robotics / driving & comma2k19~\citep{data_comma2k19} & \evPerc & \mdImage & 3{,}000 \\
    \zb Industrial acoustics & MIMII~\citep{data_mimii} & \evPerc & \mdAudio & 1{,}784 \\
    Traffic / transportation & NGSIM~\citep{data_ngsim} & \evProc & \mdTraj & 209 \\
    \zb Scientific computing & PDEBench~\citep{data_pdebench} & \evQuant & \mdField & 1{,}500 \\
    Knowledge engineering & ogbl-biokg~\citep{data_ogb} & \evSym & \mdGraph & 5{,}088{,}434 \\
    \bottomrule
  \end{tabularx}
\end{table}

\subsection{Agentic reasoning and orchestration}
\label{sec:rel-agentic}
Agentic reasoning provides the control foundation for autonomous scientific discovery.
Early methods let a single agent interleave reasoning with tool use \citep{wei2022cot, yao2023react, schick2023toolformer, li2026auto}, revise its behaviour through verbal reflection \citep{li2026taming, shinn2023reflexion}, and act on multimodal observations \citep{yang2023mmreact}.
Scientific systems extend this paradigm through specialised roles.
Agent laboratories divide the work among planning, coding, and reviewing agents \citep{schmidgall2025agentlab}, virtual laboratories coordinate teams of domain experts \citep{swanson2025virtuallab}, and collaborative ecosystems connect agents through shared knowledge and review \citep{shao2025omniscientist, li2026fortis}.
Yet these systems orchestrate agents through prompts and generated messages, which leaves stage transitions and validation dependent on fallible model output.
\OmniName adopts a \emph{pipeline of agents}, in which code controls the transitions, verifies that a stage's output is grounded in real observations, and backtracks when it is not.
Reasoning inside a stage stays open-ended, and the process around it stays predictable.

\section{Problem setting}
\label{sec:prelim}

\subsection{Scientific evidence}
\label{sec:evidence}
Scientific discovery draws on research artifacts in several forms, including images, symbolic structures, numerical results, and records of experimental processes.
Existing taxonomies typically organise scientific machine-learning data by representation, such as images, sequences, and graphs \citep{wang2023scidiscovery}, while surveys of multimodal learning adopt similar schemes \citep{baltrusaitis2019multimodal}.
To distinguish artifacts by the primary reasoning required for their interpretation, we group scientific evidence into 4 discipline-independent families (\cref{tab:evidence}): (1) the \textbf{perceptual} family, encompassing images, micrographs, spectra, waveforms, and 3-D structures; (2) the \textbf{symbolic} family, which covers evidence expressed in natural language or formal notation (e.g., documents, formulae, sequences, and knowledge graphs); (3) the \textbf{quantitative-statistical} family, comprising tables, measurements, and distributions; and (4) the \textbf{procedural} family, capturing trajectories, simulations, and agent traces. 
Here, \emph{symbolic} covers evidence expressed in natural language or formal notation.
Scientific multimodal benchmarks and domain foundation models already target many perceptual modalities \citep{yue2023mmmu, li2024mmsci, chen2024uni, parker2023astroclip, jakubik2023prithvi}.
The other 3 families account for the symbolic, statistical, and procedural artifacts that a research loop must also handle.

\begin{figure}[!t]
  \centering
  \includegraphics[width=\linewidth]{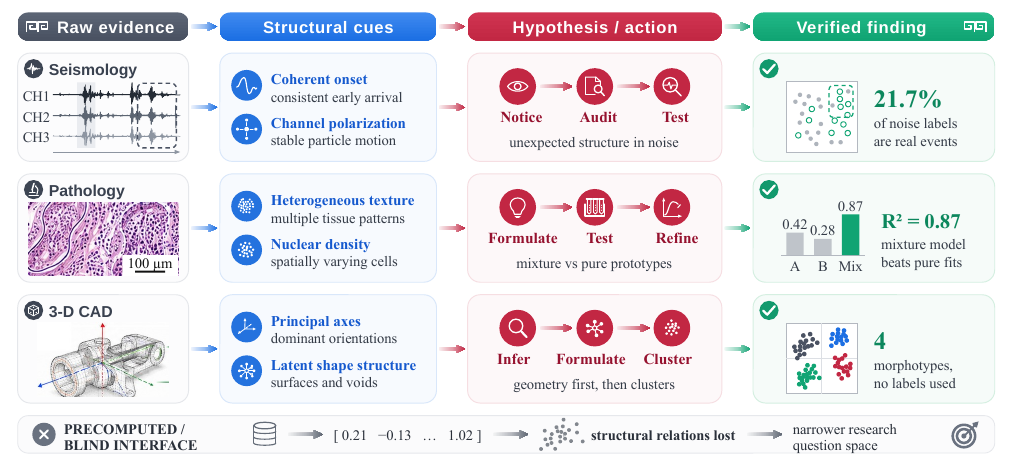}
  \caption{Progression from raw evidence to verified findings across three demonstration cases. The top three rows track workflows in seismology, pathology, and 3-D CAD. From left to right, each row begins with raw evidence (a three-channel seismogram, a stained pathology tile, and a 3-D CAD model), identifies specific structural cues, and outlines the subsequent hypothesis and action sequence. The rightmost column displays the verified findings, such as the discovery that 21.7\% of noise labels are real events. The bottom band depicts a precomputed interface where the artifact is reduced to a feature vector, resulting in lost structural relations and a narrower research question space.}
  \label{fig:interface}
\end{figure}

Current AI-scientist systems primarily process text and numerical data, which leaves perceptual and procedural evidence unexamined and narrows both the questions they can ask and the disciplines they can serve.
Both families can be serialised into tokens.
\textbf{The relevant question is not what can be serialised, but which relations survive the interface.}
Captions used as textual summaries do not preserve the local spatial structure of pathology tiles, Sentinel-2 scenes, and three-component seismograms, and unordered scalar summaries discard the temporal ordering of migration tracks and simulations.
Text-only interfaces built from these reductions therefore lose the structure on which the scientific conclusion depends.
\Cref{fig:interface} makes the contrast concrete on 3 cases of our suite, where the structural cues read off the raw record are what carry the finding, while the same artifact delivered as a precomputed vector arrives with those relations already removed.

\begin{table}[!t]
  \centering
  \caption{The 4 families of scientific evidence \OmniName is designed to
    perceive. }
  \label{tab:evidence}
  \renewcommand{\arraystretch}{1.28}\setlength{\tabcolsep}{7pt}
  \begin{tabularx}{\linewidth}{@{}l >{\raggedright\arraybackslash}X@{}}
    \toprule
    \textbf{Evidence family} & \textbf{Typical artifacts} \\
    \midrule
    \textbf{\textcolor{omniN}{Perceptual}} &
      Images, video, micrographs, radar, astronomical and remote-sensing imagery,
      the visual form of scientific plots, audio, and 3-D structure. \\
    \addlinespace[2pt]
    \textbf{\textcolor{omniI}{Symbolic}} &
      Natural-language documents, formulae, variables, rules, sequences, knowledge
      graphs, logical and causal relations, mathematical models. \\
    \addlinespace[2pt]
    \textbf{\textcolor{omniM}{Quantitative-statistical}} &
      Tables, measurements, distributions, curves, correlations, significance
      tests, regression results. \\
    \addlinespace[2pt]
    \textbf{\textcolor{omniO}{Procedural / dynamic}} &
      Experimental steps, code execution, agent traces, simulations, dynamic
      evolution, protocols. \\
    \bottomrule
  \end{tabularx}

\end{table}

\subsection{Task setting and demonstration suite}
\label{sec:prelim-suite}
A task is presented to the system as a single specification file that details a dataset, a scientific subject, and a target property, alongside the corresponding raw data.
Based on these inputs, the system is instructed to produce an evidence-grounded paper.
Nothing else is supplied, and the specific methodology is left entirely to the agent.
To evaluate this task formulation across disciplines, we assemble a demonstration suite comprising 5 top-level discipline categories and 36 second-level cases (\cref{tab:disciplines}).
Each case utilizes one real, publicly downloadable dataset with a canonical citation.
The suite ranges in scale from 12 symbolic-regression equations to a biomedical knowledge graph with 5 million edges, incorporating images, spectra, waveforms, audio, video, 3-D structures, tables, and symbolic graphs.
All 4 aforementioned evidence families are represented.
The perceptual family accounts for 28 of the 36 cases, reflecting the natural prevalence of image, signal, audio, video, and 3-D data in the selected disciplines.
The remaining 8 cases from the symbolic, quantitative-statistical, and procedural families serve as breadth controls for settings where visual inspection contributes little and a text-only baseline is already expected to perform strongly.

This comprehensive suite enables rigorous testing of the engine's domain agnosticism.
\textbf{Expanding to a new discipline requires merely writing an additional specification file.}
The underlying engine remains entirely unchanged; the identical perception, ideation, experimentation, and write-up loop operates seamlessly on a seismogram, a CAD mesh, or a knowledge graph without a single line of domain-specific code.
\Cref{app:spec} reproduces one such file in full.
Finally, \cref{tab:eval-breadth} reports the end-to-end runs enabled by this unified framework, quantitatively evaluating the true extent of the system's cross-disciplinary capabilities.

 \section{The \texorpdfstring{\OmniName}{OmniScientist} framework}
\label{sec:framework}
\begin{figure}[!t]
  \centering
  \includegraphics[width=\linewidth]{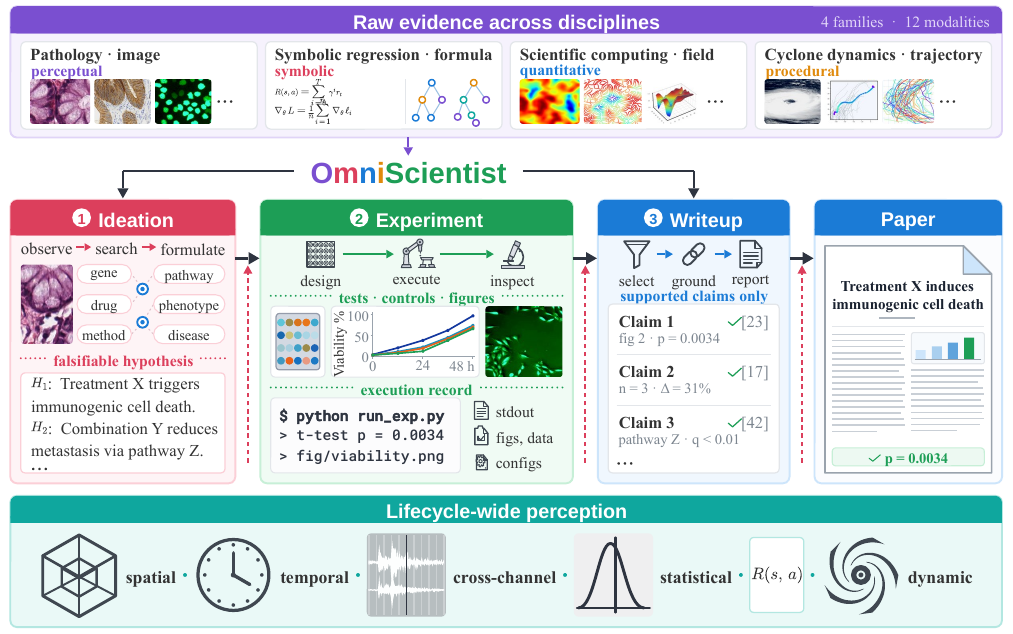}
  \caption{Architecture of the \OmniName framework. At the top, raw evidence from multiple disciplines enters the system, categorized into four evidence families (perceptual, symbolic, quantitative, and procedural) and 12 modalities. The core pipeline consists of three sequential stages. First, the Ideation stage (left) observes materials, searches literature, and formulates falsifiable hypotheses. Next, the Experiment stage (center) designs tests, executes code, and inspects results to generate an execution record containing standard output, figures, data, and configurations. Finally, the Writeup stage (right) selects, grounds, and reports claims supported exclusively by the execution record to compile the final paper. At the bottom, a lifecycle-wide perception layer provides spatial, temporal, cross-channel, statistical, and dynamic analysis capabilities. Dashed arrows indicate that these perception tools are available to all three stages of the pipeline.}
  \label{fig:arch}
\end{figure}

In this section, we introduce the \OmniName framework, an end-to-end AI scientist consisting of a perception layer and 3 autonomous agents for ideation, experiment, and writeup (\cref{fig:arch}).

\subsection{Perception layer}
\label{sec:perception}
\begin{figure}[p]
  \centering
  \includegraphics[width=\linewidth]{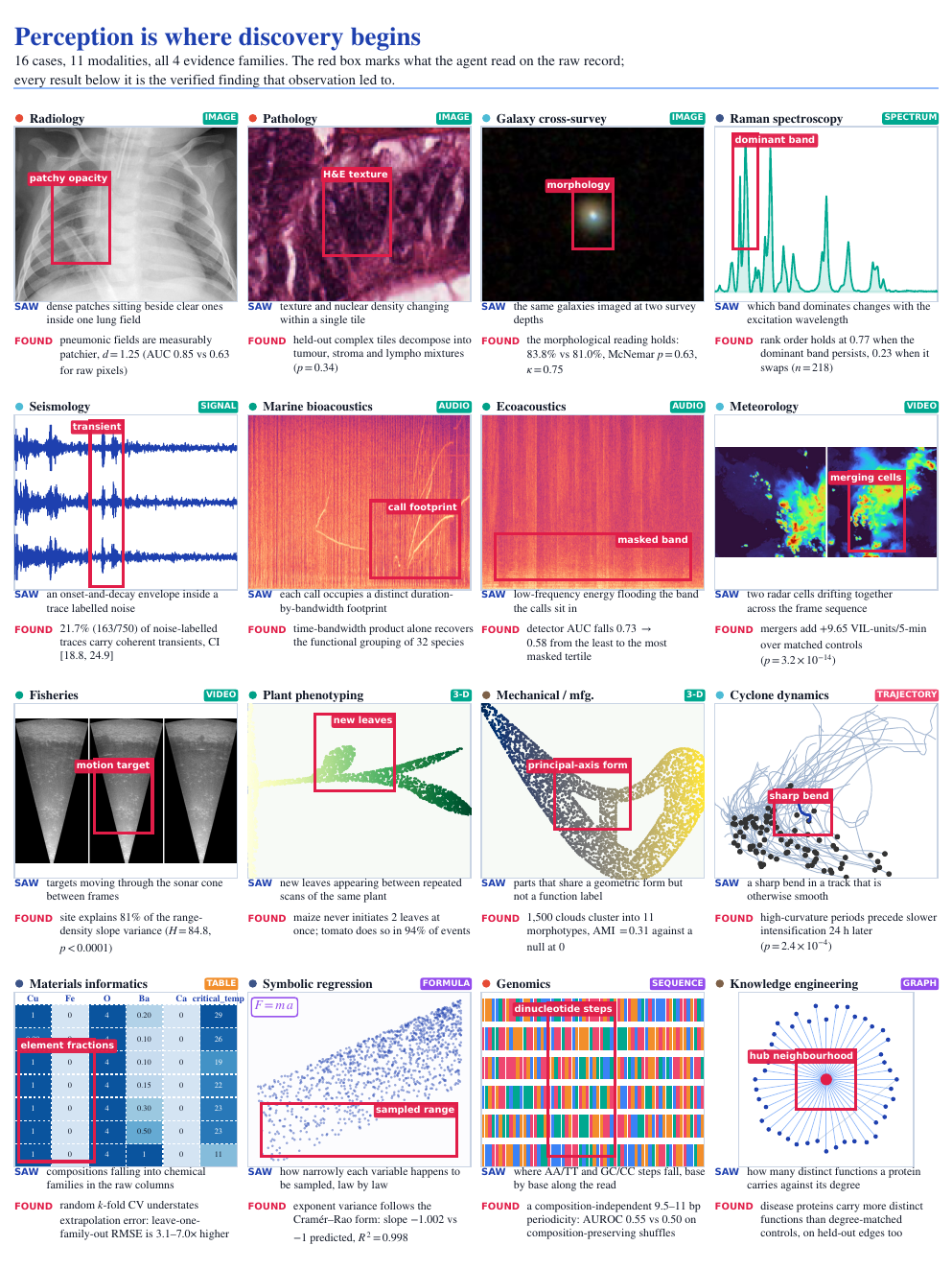}
  \caption{Raw observations and derived discoveries processed by the perception layer across 16 cases, 11 modalities, and all 4 evidence families. The figure presents a four-by-four grid of artifacts, each taken from that case's own data exactly as the run received it, with the discipline named at the top left of every panel and the modality at the top right. Within each artifact, a red bounding box marks the specific feature flagged by the agent on the raw record. Below the artifact, the \textsc{saw} label reports the direct observation made by the agent, and the \textsc{found} label details the verified experimental result produced by that observation. The first three rows cover perceptual and procedural evidence, spanning images, spectra, signals, audio, video, three-dimensional structures, and trajectories. The bottom row presents quantitative and symbolic evidence, where the layer reads the native numeric structure of a table, a formula, a sequence, or a graph instead of rendering an image.}
  \label{fig:perception-gallery}
\end{figure}

To function effectively, a multimodal AI scientist must perceive raw artifacts directly, upstream of any pre-computed summaries.
The perception layer supplies this grounding by organizing observations hierarchically.
First, artifacts are categorized into an \emph{evidence family} based on the reasoning paradigms they require.
Within a given family, a specific \emph{modality} defines the artifact's exact representation (e.g., images, tables, or time-series signals), which the agent inspects using registered tools.
\Cref{fig:perception-gallery} illustrates the raw observations processed by the agent across 16 disciplines and 11 modalities, covering all 4 evidence families, along with the discoveries derived from this direct reading.

To balance thorough analysis with computational efficiency, the framework governs \emph{when} and \emph{how} the agent inspects raw data.
Rather than immediately rendering visual plots, the agent prioritizes native numeric analysis by extracting key properties, such as FFT peaks or trend points, directly from the raw artifacts.
Visual rendering is invoked only when spatial or structural patterns are essential.
Furthermore, visual perception is budget-constrained to prevent unnecessary processing and enforce targeted inspection.
Ultimately, the task context dynamically determines whether native numeric features, visual representations, or both are utilized, ensuring flexible perception without relying on hardcoded heuristics.
\Cref{app:tools} lists the tools this suite exposed, the modality that unlocks each one, and how heavily each was used across the suite.

\subsection{Ideation}
\label{sec:ideation}
The ideation stage requires the agent to formulate a concrete, novel, and falsifiable question that can be answered computationally from the supplied data.
Driven by a ReAct~\citep{yao2023react} loop, the agent autonomously sequences its discovery process.
It begins by establishing grounding through an inventory of the materials and a decision on whether to inspect raw observations (\cref{sec:perception}).
It then contextualizes these findings by searching the literature through OpenAlex (\citealp{priem2022openalex}) with Crossref (\citealp{hendricks2020crossref}) as a fallback.
Building on this context, the agent develops at least 5 candidate ideas, assesses the novelty risk and feasibility for each, and selects the strongest candidate to finalize.

Because large language models are prone to hallucination and overconfidence, the stage output must pass a code-enforced check before it can be finalised.
The check validates structural completeness by requiring a clear research question, hypothesis, experiment sketch, and falsification criterion.
It verifies the thoroughness of the generation process by checking for the 5 self-filtered candidates and at least 3 focused literature searches.
The proposed study must be executable in code and must not require a physical experiment.
The system also requires leakage checks, effective-sample estimates, and visual audits to prevent methodological flaws.
Furthermore, since a single bounded search cannot establish absolute priority, the system automatically tempers overconfident assertions by rewriting terms like \emph{first} or \emph{never explored} to \emph{appears under-explored based on this search}.
\Cref{app:checks} states the loop this check terminates, lists every condition the 3 checks enforce, and reports which of them actually fired across the 36 runs.

\subsection{Experiment}
\label{sec:experiment}

\begin{figure}[!t]
  \centering
  \includegraphics[width=\linewidth]{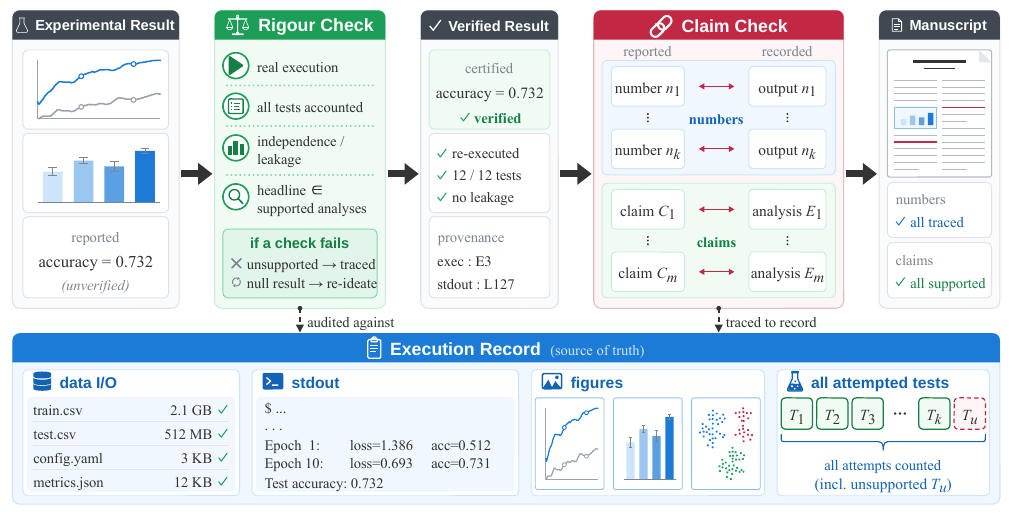}
  \caption{Two-stage verification pipeline for experimental results and manuscript claims. In the top row from left to right, an unverified experimental result undergoes a rigour check that verifies real execution, accounts for all tests, tests for independence and leakage, and ensures the headline belongs to supported analyses. If a check fails, unsupported analyses are traced and null results trigger re-ideation. Successful validation yields a verified result with certified metrics and attached provenance. Further right, a claim check matches reported numbers ($n_1 \dots n_k$) to recorded outputs and reported claims ($C_1 \dots C_m$) to recorded analyses ($E_1 \dots E_m$), resulting in a manuscript with fully traced numbers and supported claims. The bottom band displays the execution record, which serves as the source of truth for both checks. This record captures data I/O, standard output, generated figures, and a complete list of all attempted tests including unsupported attempts ($T_u$).}
  \label{fig:checks}
\end{figure}

During experimentation, the agent autonomously translates the finalized idea into a methodological design and implements it through iterative code generation.
It relies on a controlled \texttt{run\_python} environment that manages subprocess execution and figure capture.
The agent operates in a continuous debugging loop, analyzing execution errors and regenerating scripts until successful.
Throughout this process, it utilizes the perception layer to inspect raw input data or verify structural patterns in its own generated experimental plots.
To ensure findings are robust and defensible, the agent incorporates a comprehensive suite of at least 4 analyses into the experimental design, combining a main hypothesis test with essential controls such as baselines, ablation studies, mechanism probes, or sensitivity sweeps.

Once the iterative execution concludes, a code-enforced exit check verifies result provenance and statistical validity (\cref{alg:gate}).
The check grounds the experiment in reality by confirming that the agent genuinely accessed the requested dataset and generated figures matching the raw execution trace.
To prevent statistical manipulation, the system enforces a strict multiple-comparison correction that accounts for every test attempted during the debugging loop, which prevents artificial reductions in the correction denominator.
It also applies a post-hoc rescue guard to ensure headline findings originate from primary tests, runs an independence check to eliminate circular predictions, and relegates any unsupported analyses to the execution trace.
\Cref{fig:checks} places this check and the later claim check in sequence, both auditing against the same execution record rather than against the text the agent wrote.

\begin{algorithm}[t]
  \caption{Code-enforced exit check on result provenance,
    anti-fabrication, and anti-HARKing.}
  \label{alg:gate}
  \begin{algorithmic}[1]
    \Require reported result $R$, set of real stdout strings $O$
    \State \textbf{if} $R.\textrm{metric} = \emptyset$ \textbf{then reject}
    \State \textbf{if} no run in $O$ succeeded \textbf{then reject}
    \Comment{must actually execute code}
    \ForAll{number $n$ reported in $R$}
      \State \textbf{if} $n \notin O$ \textbf{then reject}
      \Comment{every number traces to real output}
    \EndFor
    \State \textbf{if} dataset was not loaded from disk \textbf{then reject}
    \If{$R$ reports $\ge 2$ $p$-values}
      \State correct over \emph{all} tests run, including demoted ones
    \EndIf
    \State \textbf{if} headline $\notin$ supported analyses \textbf{then reject}
    \Comment{anti-HARKing}
    \State keep unsupported non-headline analyses in the trace
    \State \Return \textbf{accept}
  \end{algorithmic}
\end{algorithm}

\subsection{Writeup}
\label{sec:writeup}
The writeup stage is where breadth becomes visible in the artifact itself.
A single template would make every case read like the same discipline, so the stage carries 5 structural specifications that fix the skeleton and the length of each venue style.
Machine learning papers carry Related Work and Limitations, biomedical papers append Methods at the end, and chemistry papers merge Results and Discussion.
The style is resolved from the case specification or inferred from the subject and stays decoupled from the research domain, so one engine writes a seismology study and a materials study in the idiom each field expects (\cref{app:writeup}).
Guided by this blueprint, drafting proceeds from a section-level outline into full paragraphs.
Each section is expanded only from the slice of the structured experiment record it needs, so methodological detail reaches the method and data sections, the decisive numbers reach the results section, and a section cannot introduce detail it was never given.
The abstract is written last from the same records, so its numbers agree with the main text.
The remaining machinery is deterministic.
A thesis planner selects the headline claim from the supported analyses and assigns the other results to supporting evidence, controls, or robustness checks, with the remainder kept in the run's trace.
References are retrieved through the OpenAlex API, an output pass filters the drafted sections and rolls back if it would prune too much, and a final meta-audit checks the generated claims against the experimental record (\cref{fig:checks}) before the stage compiles the PDF.
\Cref{app:prompts} gives the prompt that drives each stage verbatim, together with the rubric the judges receive.

\begin{table}[!t]
  \centering
  \renewcommand{\arraystretch}{1.00}\setlength{\tabcolsep}{2.5pt}
  \caption{
  Detailed review scores across reasoning backbones. Per-dimension means (0–10) are derived from a 2-judge cross-family panel (deepseek-v4-flash and gemini-2.5-flash-lite) over the entire case suite. For these evaluations, the framework and perception models are held fixed, with only the reasoning backbone swapped. Failed runs are excluded; thus, means are computed exclusively over successfully scored papers. The highest value in each column is highlighted, and coverage per backbone is detailed in Table 4. Notably, clarity exhibits the least degradation, whereas factual accuracy and soundness most closely track the underlying backbone strength.
  }
  \label{tab:eval-backbone}
\resizebox{\linewidth}{!}{%
  \begin{tabular}{l *{5}{c} *{2}{c} c}
    \toprule
    & \multicolumn{5}{c}{\textbf{Standard peer-review}}
    & \multicolumn{2}{c}{\textbf{MM-mandatory}} & \\
    \cmidrule(lr){2-6}\cmidrule(lr){7-8}
    \textbf{Backbone} & Novelty\hib & Sound.\hib & Clarity\hib & Signif.\hib & Reprod.\hib & MM-grnd\hib & Factual\hib & \textbf{Overall}\hib \\
    \midrule
    Sonnet~5~\citep{claude} & \best{6.3} & 7.0 & \best{7.0} & 6.3 & \best{6.1} & 5.1 & 7.7 & 6.3 \\
    \zb GPT-5.6~\citep{gpt56}    & 5.2 & 6.3 & 6.3 & 5.0 & 5.2 & 4.2 & 7.7 & 5.6 \\
    GLM-5.2~\citep{glm5}         & 6.2 & 7.1 & 6.8 & \best{6.4} & 5.9 & \best{6.6} & 7.5 & \best{6.5} \\
    \zb Kimi~K2.7~\citep{kimik2} & 6.2 & \best{7.2} & 6.7 & 6.2 & 5.5 & 5.8 & \best{8.0} & 6.2 \\
    Qwen3.5-122B~\citep{qwen3}   & 4.7 & 5.5 & 6.2 & 4.8 & 4.8 & 4.8 & 6.5 & 5.1 \\
    \zb Qwen3.5-27B~\citep{qwen3}& 5.0 & 5.6 & 5.9 & 4.9 & 4.6 & 4.9 & 6.4 & 5.1 \\
    Qwen3.5-9B~\citep{qwen3}     & 4.0 & 4.1 & 4.8 & 3.7 & 3.7 & 3.9 & 4.8 & 4.0 \\
    \zb Gemma-4-31B~\citep{gemma4}& 4.7 & 5.0 & 5.6 & 4.5 & 4.4 & 4.6 & 6.5 & 4.8 \\
    Gemma-4-26B~\citep{gemma4}   & 4.4 & 4.4 & 5.0 & 4.0 & 3.7 & 3.8 & 5.1 & 4.2 \\
    \bottomrule
  \end{tabular}}
\end{table}

\section{Evaluation}
To comprehensively evaluate our proposed framework, we conduct extensive experiments across 36 datasets to verify its effectiveness. Furthermore, we present detailed ablation studies and in-depth analyses to elucidate the underlying mechanisms and demonstrate the overall superiority of the system. We also present 2 representative case studies to show that the system can conduct interesting and useful scientific discovery.
\label{sec:eval}

\subsection{Experimental Setup}
\label{sec:setup}
\paragraph{Models} Three roles run on separate models, and keeping them apart is what makes
the comparison interpretable.
The \emph{reasoning backbone} drives all 3 stages and is the only component we swap.
We use Claude Sonnet~5~\citep{claude} for the primary runs and compare against GPT-5.6~\citep{gpt56}, GLM-5.2~\citep{glm5}, Kimi~K2.7~\citep{kimik2}, and the open-weight Qwen3.5~\citep{qwen3} (9B, 27B, 122B) and Gemma-4~\citep{gemma4} (26B, 31B) families.
The closed models are called through their official APIs, and the open-weight models are served locally.
The \emph{perception model} is pinned to Claude Sonnet~5~\citep{claude} in every run and never follows the backbone, so every look at a raw observation is served by the same model and a change in score is attributable to reasoning alone.
\emph{Scoring} is done by 2 judges from families outside the systems under test, \texttt{deepseek-v4-flash} and \texttt{gemini-2.5-flash-lite}.
\Cref{tab:eval-breadth} records how much of the suite each backbone actually completed.

\begin{table}[!t]
  \centering
  \setlength{\tabcolsep}{6pt}
  \caption{Backbone generality across the 36-case suite. The table reports the number of cases dispatched, the resulting completed papers, and the mean composite score for these successful runs.}
  \label{tab:eval-breadth}
  \begin{tabular}{@{}l*{9}{c}@{}}
    \toprule
    \textbf{Backbone} & \stkhd{Sonnet}{5} & \stkhd{GPT}{5.6} & \stkhd{GLM}{5.2}
      & \stkhd{Kimi}{K2.7} & \stkhd{Qwen3.5}{122B} & \stkhd{Qwen3.5}{27B}
      & \stkhd{Qwen3.5}{9B} & \stkhd{Gemma-4}{31B} & \stkhd{Gemma-4}{26B} \\
    \midrule
    \textbf{Cases}              & 36 & 10 & 18 & 9 & 34 & 36 & 32 & 36 & 34 \\
    \textbf{Completed}\hib      & 36 & 9 & 17 & 6 & 30 & 32 & 18 & 32 & 25 \\
    \textbf{Mean}\hib           & 6.5 & 5.7 & 6.7 & 6.5 & 5.4 & 5.3 & 4.1 & 5.0 & 4.3 \\
    \bottomrule
  \end{tabular}
\end{table}

\paragraph{Parameters}
Each stage operates under explicit computational limits to ensure bounded exploration and convergence.
The ideation stage permits a maximum of 24 agent steps and 8 literature queries.
Additionally, its visual perception is governed by an adaptive image budget of $\min(24, \max(8, 2g))$ inspections, where $g$ denotes the number of label groups in the data.
The experimentation stage runs for at most 50 steps with 8 visual inspections, and each \texttt{run\_python} execution is strictly terminated after a 150-second timeout.
To prevent infinite loops, the pipeline allows a maximum of 2 fallbacks to the ideation stage before halting the process.
Finally, text generation is capped at 8,000 tokens per call.
Under these configurations, a complete 3-stage execution costs between \$0.03 and \$4.34 depending on the backbone model, with the experimentation stage accounting for the majority of the expense.
Each complete run outputs a structured JSON record, a Markdown summary, a replayable execution trace, and a compiled PDF report.
\Cref{app:runstats} reports what a run consists of in practice, recovered from those traces.

\paragraph{Metrics}
Each generated paper is evaluated across 7 dimensions on a scale from $0$ to $10$.
These include 5 standard peer-review criteria (novelty, soundness, clarity, significance, and reproducibility), alongside 2 task-specific metrics: multimodal grounding and factual accuracy.
We report a composite score, calculated as the mean of these 7 dimensions, as well as the judges' independent overall scores.
Notably, these scores exhibit minimal correlation with manuscript length ($\rho=0.16$), indicating that the evaluation is robust against verbosity bias.
Finally, we report the system's self-assessed verdict for each hypothesis.
This verdict is extracted directly from the verified experiment record and categorized as \emph{supported}, \emph{mixed}, \emph{refuted}, or \emph{null}.
Because Sonnet~5 demonstrates the highest stability across the complete evaluation suite, we designate it as the primary reasoning backbone for all subsequent analyses.
Alternative models, such as GPT, GLM, and Kimi, were evaluated on specific subsets of the suite and are included for comparative reference.

\subsection{Main Results}
\label{sec:eval-main}
\begin{figure}[!t]
  \centering
  \includegraphics[width=\linewidth]{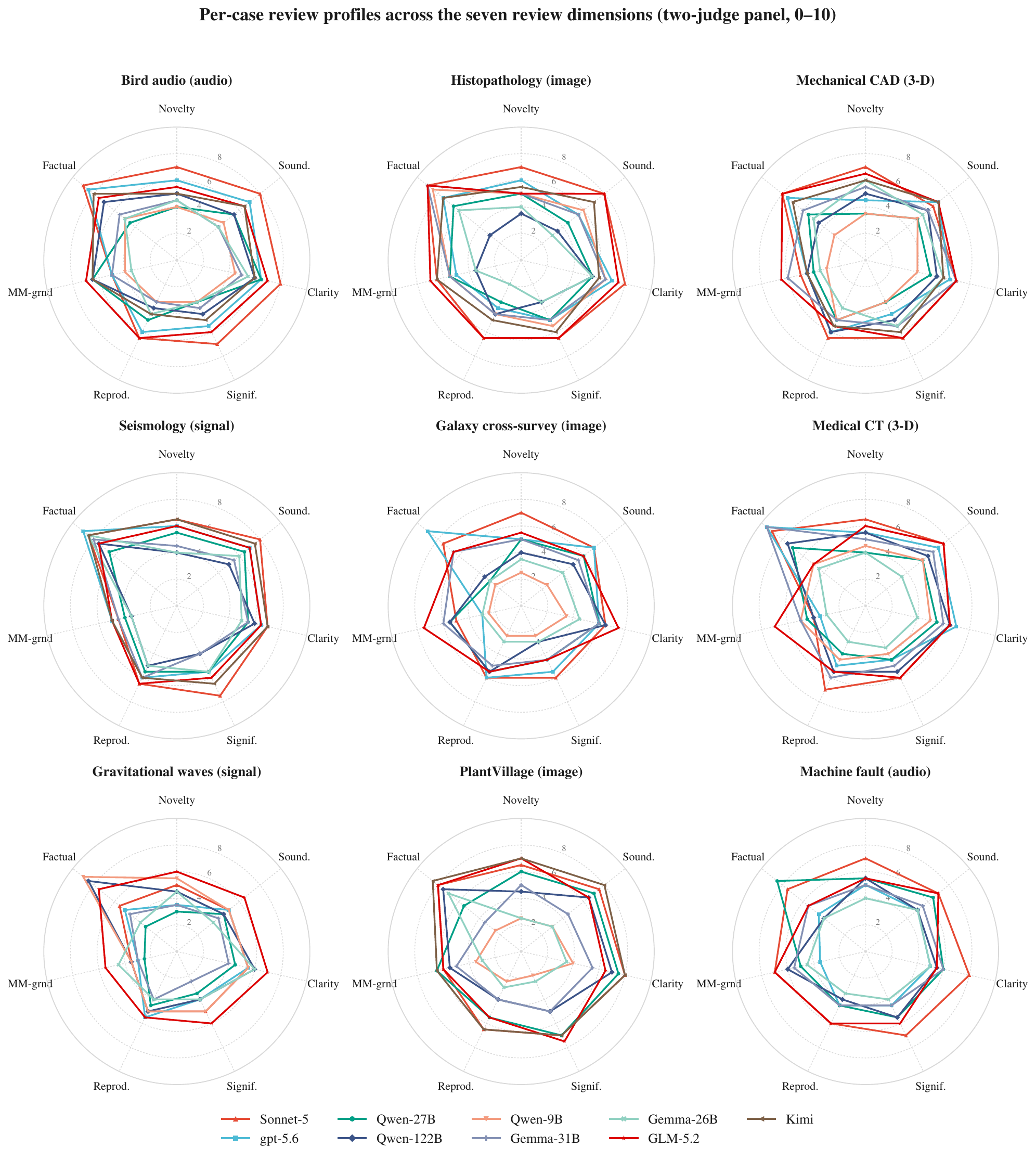}
  \caption{Per-case review profiles across the 7 dimensions. Radar plots are shown for 9 high-coverage cases spanning 4 evidence modalities; each line represents one backbone, scored by a 2-judge panel (on a $0$--$10$ scale). The strong backbones (Sonnet~5, GLM, Kimi) exhibit the largest, most balanced profiles, while the weak open models (Qwen3.5-9B, Gemma-4-26B) collapse inward, particularly in novelty and significance, although clarity varies the least across all models.}
  \label{fig:eval-radar}
\end{figure}

As shown in \cref{tab:eval-backbone} and \cref{fig:eval-radar}, we highlight 3 key findings regarding the end-to-end quality of the generated manuscripts.

First, end-to-end generation achieves high quality and is consistent across top LLM backbones. \OmniName automates the entire research pipeline and returns complete manuscripts across the suite. Powered by Claude, the framework achieves an overall score of $6.3$ on the full suite. The strongest alternate backbones, such as GLM and Kimi, fall within a remarkably similar performance range on their respective evaluated subsets.

Second, performance is highly robust across all disciplines and evidence modalities. Across the evaluated cases, generation quality remains remarkably stable regardless of the research domain or data type. Whether grouping the runs by discipline or by evidence modality, the median composite scores range tightly between $6.1$ and $7.1$ (\cref{tab:eval-rubric-suite,tab:eval-aggregate}). Furthermore, the highest-scoring manuscripts broadly span all domain categories, demonstrating strong generalization.

Third, factual accuracy consistently leads the evaluation metrics across all reasoning engines. The relative ordering of the qualitative evaluation dimensions is highly concordant across the various backbones, achieving a median pairwise Spearman correlation of $0.82$. Across these models, factual accuracy consistently ranks at the top. This uniformity aligns directly with the shared provenance requirement, which holds each reasoning engine to the same rigorous evidence standards, while exit checks anchor every reported outcome to real program output regardless of model fluency.

\begin{table}[!t]
  \centering
  \setlength{\tabcolsep}{3pt}\renewcommand{\arraystretch}{1.0}\setlength{\extrarowheight}{0pt}
  \caption{Backbone quality aggregated by evidence modality and discipline family. Results are grouped by modality on the left and by discipline family on the right. The reported metric is the mean composite score evaluated by the 2-judge cross-family panel on a scale of $0$ to $10$. A dash indicates that a backbone produced no scored papers for that specific category.}
  \label{tab:eval-aggregate}
  \begin{tabularx}{\linewidth}{@{}l *{7}{Y} !{\hspace{3pt}\color{nusgray!55}\vrule\hspace{3pt}} *{5}{Y}@{}}
    \toprule
    & \multicolumn{7}{c}{\textbf{By evidence modality}} & \multicolumn{5}{c}{\textbf{By discipline}} \\
    \cmidrule(lr){2-8}\cmidrule(lr){9-13}
    \textbf{Backbone} & \textbf{Image} & \textbf{Signal} & \textbf{Audio} & \textbf{Video} & \textbf{3-D} & \textbf{Traj.} & \textbf{T\&S} & \textbf{Earth} & \textbf{Life} & \textbf{Agri.} & \textbf{Engin.} & \textbf{Phys.} \\
    \midrule
    Sonnet~5       & 6.4 & 6.1 & \best{7.1} & \best{6.4} & \best{7.0} & \best{6.3} & \best{6.4} & 6.5 & 6.7 & \best{6.8} & \best{6.6} & 5.8 \\
    \zb GPT-5.6    & 5.9 & 5.5 & 5.5 & -- & 5.9 & -- & -- & 5.6 & 6.0 & 6.4 & 5.2 & -- \\
    Qwen3.5-27B    & 5.5 & 5.0 & 5.9 & 5.3 & 4.8 & 4.9 & 5.8 & 5.1 & 5.5 & 5.5 & 4.8 & 5.5 \\
    \zb Qwen3.5-9B & 3.8 & 5.5 & 4.1 & 4.5 & 4.4 & 4.3 & 3.2 & 4.1 & 4.5 & 4.0 & 4.4 & 3.2 \\
    Qwen3.5-122B   & 4.8 & 5.4 & 5.9 & 5.2 & 5.4 & 5.8 & 5.5 & 4.6 & 5.7 & 6.0 & 4.9 & 5.4 \\
    \zb Gemma-4-31B& 5.3 & 4.9 & 5.4 & 5.1 & 5.9 & 4.0 & 3.5 & 5.1 & 5.7 & 4.6 & 5.0 & 4.9 \\
    Gemma-4-26B    & 4.1 & 4.3 & 4.4 & 5.0 & 4.3 & 4.3 & 4.8 & 4.5 & 3.8 & 4.3 & 4.4 & 4.6 \\
    \zb GLM-5.2    & 6.6 & \best{6.8} & 6.6 & -- & 6.6 & -- & -- & 6.4 & \best{6.9} & 6.5 & 6.4 & \best{7.5} \\
    Kimi~K2.7      & \best{6.8} & 6.7 & 5.9 & -- & 6.0 & -- & -- & \best{6.8} & 6.2 & 6.6 & 6.0 & -- \\
    \bottomrule
  \end{tabularx}
\end{table}

\begin{table}[t]
  \centering \setlength{\tabcolsep}{5pt}\renewcommand{\arraystretch}{1.0}\setlength{\extrarowheight}{0pt}
  \caption{Review rubric performance across the complete evaluation suite using the Sonnet~5 backbone. The 2-judge cross-family panel evaluated all cases on a scale of $0$ to $10$. Using a single backbone ensures direct comparability across all columns. \emph{Factual accuracy} reaches $7.0$ or higher in 30 of the 36 completed cases.}
  \label{tab:eval-rubric-suite}
  \begin{tabularx}{\linewidth}{@{}l *{5}{Y} YY Y@{}}
    \toprule
    & \multicolumn{5}{c}{\textbf{Standard peer-review}}
    & \multicolumn{2}{c}{\textbf{MM-mandatory}} & \\
    \cmidrule(lr){2-6}\cmidrule(lr){7-8}
    \textbf{Discipline} & Novelty\hib & Sound.\hib & Clarity\hib & Signif.\hib & Reprod.\hib & \mbox{MM-gr.\hib} & Factual\hib & \textbf{Overall}\hib \\
    \midrule
    \grpspanA{9}{Physical sciences}
    Condensed matter & 2.5 & 2.5 & 4.0 & 2.0 & 2.0 & 4.5 & 1.0 & 2.5 \\
    \zb Vibrational spectroscopy & 6.0 & 8.0 & 7.5 & 7.0 & 7.0 & 6.0 & 9.0 & 7.0 \\
    Materials informatics & 7.0 & 7.0 & 7.5 & 8.0 & 7.0 & 4.5 & 7.0 & 7.0 \\
    \zb Molecular chemistry & 4.5 & 5.5 & 6.0 & 3.5 & 6.0 & 4.0 & 5.0 & 4.5 \\
    Symbolic regression & 7.0 & 8.0 & 7.5 & 7.5 & 7.5 & 3.5 & 9.0 & 7.0 \\
    \grpspan{9}{Earth \& space}
    Remote sensing & 7.0 & 7.0 & 6.5 & 7.0 & 6.0 & 5.5 & 7.0 & 6.5 \\
    \zb Galaxy morphology & 6.5 & 6.5 & 7.0 & 6.0 & 6.0 & 5.5 & 7.5 & 6.5 \\
    Galaxy cross-survey & 7.0 & 7.0 & 6.5 & 6.0 & 6.0 & 5.0 & 7.5 & 6.5 \\
    \zb Gravitational waves & 5.0 & 5.0 & 5.5 & 5.0 & 5.0 & 3.5 & 5.5 & 4.5 \\
    Seismology & 6.5 & 8.0 & 7.0 & 7.5 & 6.5 & 4.5 & 8.5 & 7.0 \\
    \zb Marine biology & 6.0 & 8.5 & 7.5 & 7.0 & 6.5 & 5.5 & 9.5 & 7.0 \\
    Geology / petrophysics & 6.5 & 7.5 & 8.0 & 7.5 & 7.5 & 6.0 & 9.5 & 7.5 \\
    \zb Cyclone dynamics & 6.0 & 6.5 & 7.0 & 6.0 & 5.5 & 4.5 & 7.0 & 6.0 \\
    Meteorology & 6.0 & 7.0 & 8.0 & 6.0 & 5.0 & 4.5 & 8.0 & 6.5 \\
    \grpspan{9}{Life \& medical}
    Pathology & 7.0 & 8.0 & 8.0 & 6.5 & 6.5 & 6.5 & 9.0 & 7.0 \\
    \zb Radiology & 7.0 & 7.5 & 8.0 & 6.5 & 6.5 & 6.5 & 8.5 & 7.0 \\
    Medical imaging (CT) & 6.5 & 7.5 & 6.5 & 6.0 & 7.0 & 4.0 & 9.0 & 6.0 \\
    \zb Cardiology & 6.5 & 8.0 & 8.0 & 7.5 & 6.5 & 4.5 & 9.0 & 7.0 \\
    Sleep neuroscience & 6.5 & 4.5 & 7.0 & 6.0 & 5.5 & 3.5 & 4.0 & 4.5 \\
    \zb Genomics & 7.0 & 7.0 & 7.0 & 6.0 & 6.0 & 5.0 & 7.5 & 6.5 \\
    Cell biology & 7.0 & 6.0 & 5.5 & 6.5 & 5.5 & 4.0 & 6.0 & 5.5 \\
    \grpspan{9}{Agricultural \& ecological}
    Plant pathology & 6.5 & 7.5 & 8.0 & 7.0 & 6.5 & 6.0 & 8.0 & 7.0 \\
    \zb Precision agriculture & 6.0 & 6.5 & 6.5 & 6.0 & 6.0 & 5.0 & 8.0 & 6.0 \\
    Animal behavior & 6.5 & 7.5 & 7.0 & 5.5 & 5.5 & 4.5 & 8.0 & 6.5 \\
    \zb Ecoacoustics & 7.0 & 8.0 & 8.0 & 7.0 & 6.5 & 5.0 & 9.0 & 7.5 \\
    Marine bioacoustics & 6.5 & 8.0 & 7.5 & 6.5 & 5.5 & 6.5 & 9.0 & 7.0 \\
    \zb Plant phenotyping & 6.5 & 7.0 & 8.0 & 7.0 & 6.5 & 7.5 & 8.0 & 6.5 \\
    Fisheries & 6.5 & 8.0 & 7.5 & 7.0 & 6.5 & 7.5 & 9.0 & 7.0 \\
    \zb Movement ecology & 5.5 & 6.0 & 6.5 & 5.5 & 5.5 & 5.0 & 6.5 & 6.0 \\
    \grpspan{9}{Engineering \& information}
    Mechanical / mfg. & 7.0 & 6.5 & 7.0 & 6.5 & 6.5 & 5.0 & 8.0 & 6.5 \\
    \zb Civil / surveying & 6.5 & 7.5 & 7.5 & 6.5 & 7.5 & 6.0 & 8.5 & 7.0 \\
    Robotics / driving & 6.0 & 5.5 & 5.5 & 5.0 & 4.5 & 3.5 & 7.0 & 5.5 \\
    \zb Industrial acoustics & 7.0 & 7.0 & 8.0 & 7.0 & 6.0 & 7.0 & 7.5 & 7.0 \\
    Traffic / transport & 6.5 & 8.0 & 8.0 & 6.5 & 6.0 & 4.0 & 9.5 & 6.5 \\
    \zb Scientific computing & 7.0 & 7.0 & 6.5 & 6.5 & 5.5 & 5.0 & 8.5 & 6.5 \\
    Knowledge engineering & 5.5 & 8.0 & 6.0 & 6.5 & 7.0 & 4.0 & 9.0 & 6.5 \\
    \midrule
    \textbf{All-discipline mean} & \textbf{6.3} & \textbf{7.0} & \textbf{7.0} & \textbf{6.3} & \textbf{6.1} & \textbf{5.1} & \textbf{7.7} & \textbf{6.3} \\
    \bottomrule
  \end{tabularx}
\end{table}

\subsection{The Perception and Component Ablation}
\label{sec:eval-ablation}
To evaluate the contribution of direct observation to the research process, we compare our framework against a \emph{blind} baseline.
This baseline receives only precomputed scalar features, simulating the interface of a conventional text-only system.
A cross-family panel of judges evaluates the two manuscripts head-to-head.
To ensure a more sensitive assessment than isolated scoring, we randomise the presentation order to eliminate positional bias.
Across the 5 cases featuring a scalar-blind counterpart, the perception-enabled framework consistently outperforms, winning $85\%$ of the comparisons averaged over cases.

\begin{figure}[!t]
  \begin{minipage}[t]{0.485\linewidth}
    \centering
    \includegraphics[width=\linewidth]{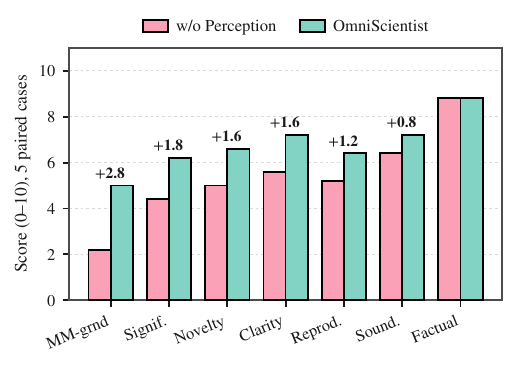}
    \caption{Dimension-wise perception gain. For the 5 cases evaluated under both conditions, the chart shows the mean scores with perception removed (pink) and for the full \OmniName (teal), scored by the same judge across both settings. The 2 panels of this row share one colour key. The largest gain is observed in multimodal grounding, while factual accuracy remains identical since both conditions undergo the same provenance check.}
    \label{fig:eval-gain}
  \end{minipage}\hfill
  \begin{minipage}[t]{0.485\linewidth}
    \centering
    \includegraphics[width=\linewidth]{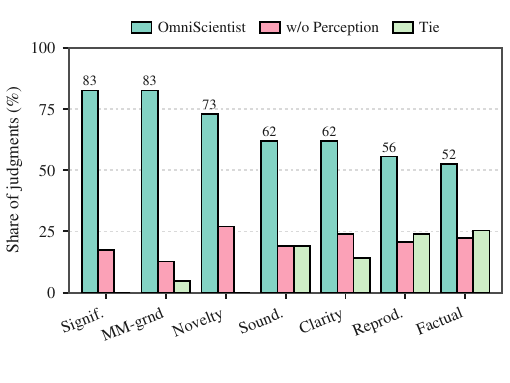}
    \caption{Breakdown of head-to-head judgments. For each dimension, the 3 bars show the share of all judgments won by \OmniName, won by the same system with perception removed, and declared a tie. Judges never tie on novelty or significance, the dimensions enhanced by perception, and tie most often on factual accuracy and reproducibility, which both conditions share through the provenance check.}
    \label{fig:eval-ties}
  \end{minipage}
\end{figure}

A dimensional analysis reveals how multimodal perception elevates manuscript quality.
When scored by the judge common to both conditions, the perception module yields the most substantial gains in multimodal grounding ($+2.8$) and significance ($+1.8$), while factual accuracy is equally high in both conditions (\cref{fig:eval-gain}).
The improvement in grounding stems from the framework's ability to process raw observations directly, a capability absent from text-only baselines.
More importantly, the gains in significance show that multimodal perception broadens the scope of the claims a study can support.
Crucially, these perceptual enhancements do not compromise empirical rigor.
Because both conditions are subject to the same strict provenance verification, every reported metric must be directly traceable to actual program outputs, regardless of the input modality.

\begin{figure}[!t]
  \begin{minipage}[t]{0.485\linewidth}
    \centering
    \includegraphics[width=\linewidth]{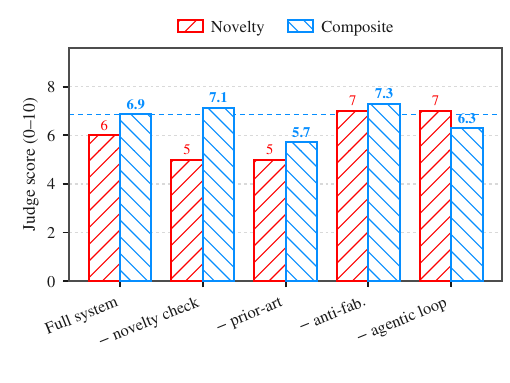}
    \caption{Component ablation on the seismology case, where each configuration removes a single component with the backbone fixed. \emph{Novelty} is the judged novelty score and \emph{Composite} the 7-dimension mean, both evaluated by the DS-V4-Flash judge ($0$--$10$). The dashed line marks the composite score of the full system.}
    \label{fig:eval-ablate}
  \end{minipage}\hfill
  \begin{minipage}[t]{0.485\linewidth}
    \centering
    \includegraphics[width=\linewidth]{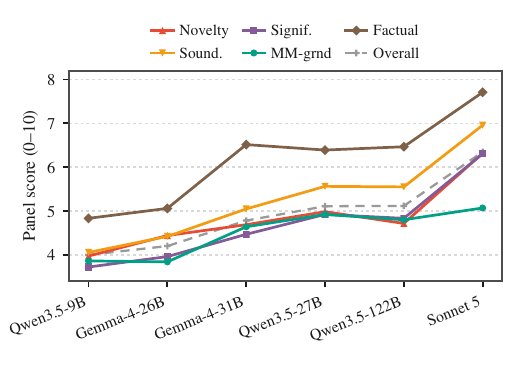}
    \caption{Review dimension scores across different backbone strengths. Data reflects the 6 backbones with the broadest case coverage, ordered by overall score. The score gap between the strongest and weakest backbones is largest for factual accuracy and smallest for multimodal grounding.}
    \label{fig:eval-dimtrend}
  \end{minipage}
\end{figure}

We conduct a leave-one-out ablation study to isolate the contributions of individual framework components (\cref{fig:eval-ablate}).
The prior-art search and the iterative agentic loop emerge as the most critical drivers of overall quality.
Omitting the prior-art search results in the steepest performance decline (from $6.9$ to $5.7$), because the system becomes prone to proposing redundant or previously published ideas.
Similarly, reducing the iterative agentic loop to a single pass significantly degrades manuscript quality.
The novelty check successfully fulfills its targeted role; its removal results in a full-point decrease in the evaluated novelty score.
Finally, provenance enforcement operates at a fundamentally different level.
Because it is a deterministic constraint enforced at the code level, it does not directly alter the narrative text evaluated by the scoring rubric.
Instead, its critical function is to mathematically guarantee the strict traceability of every reported empirical claim.

\subsection{Mechanism Analysis}
\label{sec:eval-mechanism}

\textbf{Direct observation fundamentally alters the research trajectory.}
An in-depth qualitative analysis of the 5 paired runs reveals how the performance gains discussed in \cref{sec:eval-ablation} manifest within the practical workflow.
In every instance, the perception-driven system anchors its research questions on attributes exclusive to the raw multimodal records.
Specific examples include the morphology extracted by a vision model from a galaxy image, the polarization of a three-component waveform, the texture of a pathology tile, the geometry of a CAD point cloud, and the per-point organ labels of a repeated plant scan.
Conversely, the blind baseline invariably restricts its hypothesis formulation to precomputed scalar features.
Consequently, the two systems formulate and execute entirely distinct research trajectories, demonstrating that the advantage of multimodal perception extends far beyond mere stylistic differences in manuscript composition. \Cref{tab:mechanism} details these operational divergences.
The seismology case starkly illustrates the operational deficit of the blind baseline.
Relying solely on textual feature names, the blind system designed a study that required data fields absent from the actual recording, necessitating extensive and reactive re-planning.
In contrast, the perception-driven framework leveraged direct observation to formulate a viable, immediately executable hypothesis from the outset.

\begin{table}[ht]
  \centering
  \setlength{\tabcolsep}{4pt}\renewcommand{\arraystretch}{1.12}
  \caption{
  Comparison of feature utilization between the two systems. In all paired cases, the perceiving system focuses on information inherent in the raw records, whereas the blind system relies solely on the provided scalar features despite sharing the same task and backbone.
    }
  \label{tab:mechanism}
  \begin{tabularx}{\linewidth}{@{}>{\raggedright\arraybackslash}p{2.3cm} >{\raggedright\arraybackslash}p{3.3cm} X@{}}
    \toprule
    \textbf{Case} & \textbf{Evidence only the raw record carries} & \textbf{Question each system asked} \\
    \midrule
    Galaxy cross-survey & Morphology read off the image & \emph{With perception:} does a vision model's morphological reading degrade on the shallower survey for the same galaxies? \newline \emph{Blind:} can the 3 classes be separated along the 8 supplied feature axes? \\
    \zb Seismology & Cross-component polarisation of the waveform & \emph{With perception:} what fraction of noise-labelled traces carry coherent polarised transients? \newline \emph{Blind:} do frequency-shape features retain a depth imprint after an attenuation correction? \\
    Pathology & Texture and nuclear density of the tile & \emph{With perception:} is the \textsc{complex} class a compositional mixture of the pure tissue prototypes? \newline \emph{Blind:} does the class confusion matrix follow an a priori similarity ranking? \\
    \zb Mechanical CAD & Principal-axis geometry of the point cloud & \emph{With perception:} do the function-defined labels correspond to latent geometric morphotypes? \newline \emph{Blind:} do dimension-standardised part families show tighter descriptor dispersion? \\
    Plant phenotyping & Per-point organ labels across repeated scans & \emph{With perception:} do the two species differ in how many leaves grow at once? \newline \emph{Blind:} do the species separate on shape descriptors once the size axis is removed? \\
    \bottomrule
  \end{tabularx}
\end{table}

\textbf{Perception dominates across all dimensions, with tie distributions highlighting its specific contributions.}
The head-to-head comparison also records tied outcomes, revealing a clear separation among the evaluation dimensions.
Excluding ties, the perception-driven system secures \textbf{70\% to 87\%} of the winning preferences across all 7 metrics.
The primary divergence between dimensions lies in the frequency of these tied judgments.
For multimodal grounding, significance, and novelty, the tie rate approaches zero, indicating that the two manuscripts remain consistently distinguishable in these areas.
Conversely, roughly a quarter of the comparisons result in ties for factual accuracy and reproducibility.
This consistent baseline aligns seamlessly with our design expectations, because both conditions are subject to the same strict provenance verification and only report metrics directly traceable to actual program executions.
This distinct distribution of ties is fully illustrated in \cref{fig:eval-ties}.

\begin{table}[t]
  \centering
  \setlength{\tabcolsep}{4pt}\renewcommand{\arraystretch}{1.08}
  \caption{Overview of 13 end-to-end runs, grouped by evidence modality and spanning all 4 families. Evaluation metrics are system-specific, extracted verbatim from verified experiment records.}
  \label{tab:results}
  \begin{tabularx}{\linewidth}{@{}p{3.6cm}p{1.4cm}p{1.7cm}X@{}}
    \toprule
    \textbf{Discipline} & \textbf{Evidence} & \textbf{Evaluation metric} & \textbf{Headline finding} \\
    \midrule
    Radiology         & image  & supported & Pneumonic pediatric lung fields show markedly higher local-entropy heterogeneity (patchiness) than normal. \\
    \zb Pathology     & image  & supported & The COMPLEX H\&E class is heterogeneous, splitting into compositional sub-clusters. \\
    Galaxy morphology & image  & mixed     & VLM morphology accuracy 83.8\% (DECaLS) vs 81.0\% (SDSS); the 2.8-pt gap is not significant. \\
    \zb Remote sensing& image  & mixed     & Color-only features recover 76.2\% of 10-class accuracy vs 83.2\% combined, revealing a color shortcut. \\
    Seismology        & signal & mixed     & 21.7\% of noise-labelled STEAD traces carry coherent transient bursts; the instrument-type hypothesis is refuted. \\
    \zb Cardiology    & audio  & supported & Recording-protocol metadata alone predicts abnormality (AUC 0.60) and collapses out-of-source (0.35), exposing a confound. \\
    Ecoacoustics      & audio  & supported & A mid/high-band bird-presence classifier shows a large, robust drop in discriminability across recording sets. \\
    \zb Mechanical CAD& 3-D    & supported & Scale-invariant shape descriptors cluster 1{,}500 CAD parts into function-agnostic form families without supervision. \\
    Plant phenotyping & 3-D    & mixed     & Maize initiates leaves sequentially where tomato is bursty, separable in 3-D scans. \\
    \zb Materials informatics & table & mixed & Random $k$-fold CV underestimates extrapolation error; leave-one-family-out RMSE is 3.1--7.0$\times$ higher. \\
    Symbolic regression & formula & supported & The Cram\'{e}r--Rao form $\mathrm{Var}(\hat{a})=\sigma^{2}/(N\,\mathrm{Var}\log x)$ predicts empirical exponent-estimation variance across all 8 monomial Feynman laws, sampled ranges, and noise levels. \\
    \zb Knowledge engineering & graph & supported & Disease-associated proteins carry more distinct GO-function annotations than degree-matched controls, and the excess grows with PPI degree; it replicates on withheld test-split edges. \\
    CS/ML methodology & trace  & refuted   & Rejection-driven repairs are not dominated by omission, contrary to the pre-registered hypothesis. \\
    \bottomrule
  \end{tabularx}
\end{table}

\begin{table}[!t]
  \centering
  \caption{Numbers the system produced for the STEAD noise audit. }
  \label{tab:stead}
  \begin{tabular}{@{}lc@{\hspace{1.8em}}lc@{}}
    \toprule
    \multicolumn{2}{@{}l}{\textbf{\textcolor{tblBlue}{Headline and controls}}} &
    \multicolumn{2}{l@{}}{\textbf{\textcolor{tblBlue}{Heterogeneity and robustness}}} \\
    \cmidrule(r{1.8em}){1-2}\cmidrule{3-4}
    Full-detector prevalence           & 21.7\% (163/750) & Channel BH / HH / HN prevalence   & 32.8 / 29.1 / 2.4\%   \\
    \quad 95\% confidence interval     & [18.8, 24.9]\%   & \quad channel $\chi^2$            & $p=5.7\times10^{-14}$ \\
    Amplitude-only baseline            & 2.0\%            & Per-network prevalence range      & 0--65\%               \\
    Ablation, no coincidence term      & 2.0\%            & \quad network $\chi^2$            & $p=3.5\times10^{-14}$ \\
    Null false-alarm rate (target 1\%) & 1.07\%           & Station-cluster bootstrap 95\% CI & [17.9, 25.8]\%        \\
                                       &                  & Sensitivity (FAR, null, window)   & \textcolor{nuswarm}{stable 19--25\%} \\
    \bottomrule
  \end{tabular}
\end{table}

\textbf{Robustness across disciplines and modalities.}
An examination of the aggregate results in \cref{tab:eval-aggregate} reveals that performance differences across diverse discipline families and evidence modalities become negligible once the backbone model is held constant.
Statistically, none of these cross-domain variations reach significance under a case-level permutation test, demonstrating the broad applicability of the perception-driven framework. \Cref{tab:results} lists 13 of these end-to-end runs with the system's own evaluation metric and headline finding, taken verbatim from the verified experiment record.

\textbf{Varying sensitivity to backbone scaling.}
Crucially, the evaluation dimensions exhibit varied sensitivities to backbone model capabilities.
Factual accuracy and soundness most sharply differentiate the strong foundational models from the smaller open-weight alternatives, each demonstrating a $2.9$-point performance gap between the largest and smallest models (\cref{fig:eval-dimtrend}).
In contrast, multimodal grounding is the least sensitive metric, shifting by only $1.2$ points across the identical model range.
Consequently, a stronger reasoning backbone does not inherently provide the perceptual competence that these multimodal tasks demand.

\subsection{Case Study 1: Auditing a Seismic Benchmark}
\label{sec:casestudy}
\begin{figure}[!t]
  \centering
  \includegraphics[width=0.92\linewidth]{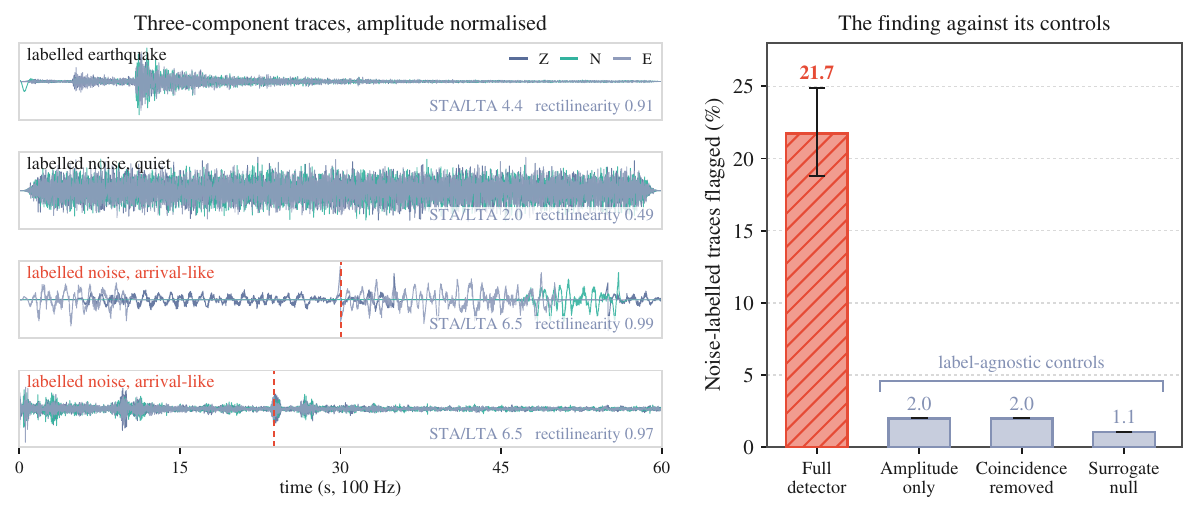}
  \caption{The seismic audit at a glance. Left: 4 of the three-component
    traces the agent read, amplitude normalised. The first is a labelled
    earthquake, shown for reference, and the second a labelled noise trace that
    really is stationary background. The last two are also labelled noise, yet
    each carries a coherent onset at the dashed line, with an STA/LTA peak above
    the upper quartile of the labelled earthquakes and an onset rectilinearity
    near $1$. All 4 were selected by the run's own stored onset statistics
    rather than by eye. Right: the share of noise-labelled traces the full
    detector flags, with its $95\%$ confidence interval, against 3
    label-agnostic controls. Removing the cross-channel coincidence term
    collapses the detector to the amplitude-only rate, which is what identifies
    timing coincidence across components as the mechanism it is using.}
\label{fig:case-stead}
\end{figure}

To investigate how direct multimodal observation drives autonomous hypothesis generation, we trace the framework's execution on auditing roughly 1{,}500 three-component broadband seismograms from the STEAD catalogue.
The input data is evenly split between earthquake and noise labels.
Upon visually processing the raw waveforms, the agent identified a clear onset-and-decay envelope within a trace explicitly labelled as noise, where only a stationary background was expected (\cref{fig:case-stead}).
The system did not override the dataset label, and instead used this conflict to formulate a targeted research question: what exact fraction of the noise-labelled traces carries coherent transient energy.
This hypothesis stems entirely from processing the raw visual morphology of the waveform, making it completely imperceptible to a baseline system that relies solely on predefined scalar features.

To resolve this formulated question, the agent autonomously engineered a composite detector integrating an STA/LTA characteristic function with amplitude, rectilinearity, planarity, and a cross-channel coincidence term.
By setting thresholds at the 99th percentile of 3 label-agnostic surrogate nulls, the system established that $21.7\%$ ($163$ of $750$) of the noise-labelled traces carry coherent, polarised, cross-component transient bursts, bounded by a $95\%$ confidence interval of $[18.8, 24.9]$ (\cref{tab:stead,fig:case-stead}).
Through an autonomous ablation study, the agent demonstrated that removing the coincidence term collapses the detection rate to a $2.0\%$ amplitude-only baseline, effectively isolating timing coincidence as the primary mechanistic driver (\cref{fig:case-stead}).
The framework verified the robustness of the $19\%$ to $25\%$ estimate across a 50-fold range of false-alarm rates, 3 distinct null models, 4 windowing choices, and a station-cluster bootstrap over 417 distinct stations.
Furthermore, when the empirical data refuted a pre-registered hypothesis concerning instrument types, the system objectively headlined the prevalence finding and relegated the instrument question to a boundary result.
Throughout this derivation, the provenance check ensured that every reported threshold and statistical claim originated directly from executable code output.
This estimate remained stable across alternative false-alarm rates, null models, windowing choices, and a station-cluster bootstrap over 417 distinct stations.

\subsection{Case Study 2: A Texture Signature in Paediatric Chest Radiographs}
\label{sec:casestudy-cxr}

\begin{figure}[!t]
  \centering
  \includegraphics[width=0.92\linewidth]{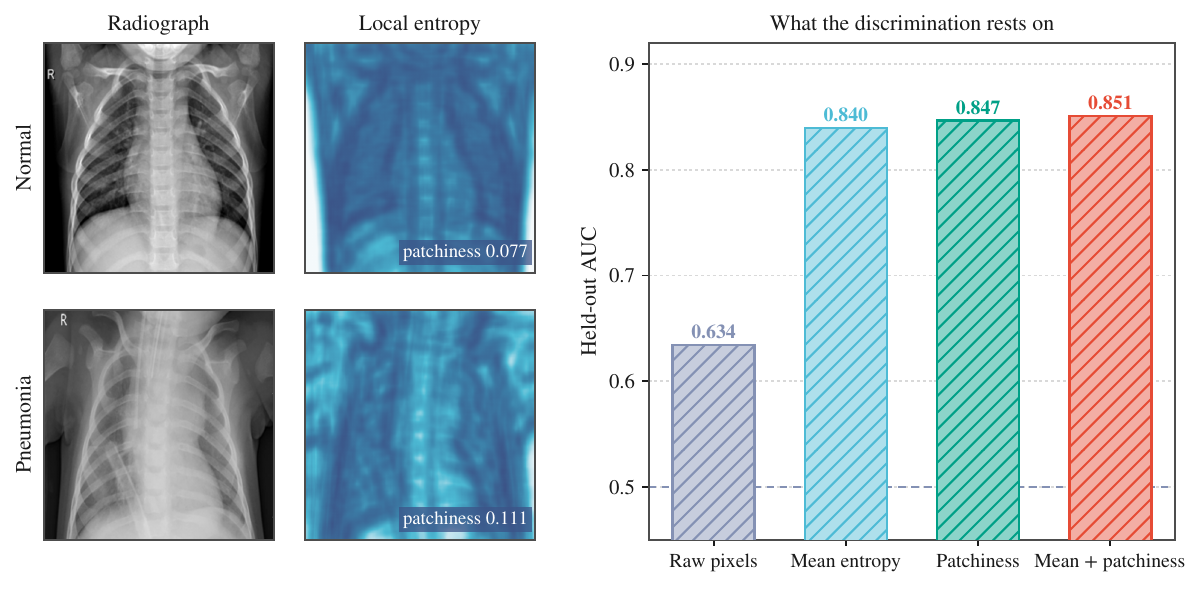}
  \caption{Visualization and evaluation of sliding-window local-entropy features. Left: Normal and pneumonic radiographs paired with their local-entropy maps. The displayed samples correspond to the median patchiness of each class. Notably, the pneumonic entropy map exhibits higher variance (i.e., a mottled appearance) compared to the relatively smooth normal map. Right: Classification performance on the held-out test set. Entropy-derived features achieve score of $0.840$--$0.851$, significantly outperforming the raw-pixel baseline ($0.634$). This indicates that the discriminative signal relies on local structural patterns rather than raw pixel intensities.}
\label{fig:case-cxr}
\end{figure}

While the previous case study demonstrated the system's ability to correct dataset labels, this second evaluation illustrates its capacity to establish a novel positive finding directly from raw visual data.
The agent received paediatric frontal chest radiographs carrying only a normal or pneumonia label, with no precomputed features.
Reading the radiographs directly, it observed that pneumonic lung fields were not uniformly brighter but unevenly mottled, with dense patches sitting beside clear ones (\cref{fig:case-cxr}).
It converted that observation into a measurable quantity, the spatial dispersion of a sliding-window local Shannon-entropy map, and pre-specified the hypothesis that this patchiness separates the two labels independently of the overall entropy level.

The observed separation is substantial and generalises robustly to out-of-sample data.
Spatial heterogeneity separates normal from pneumonic lung fields with a large effect size (Cohen's $d > 1.2$) that remains remarkably stable across both development and held-out splits (\cref{tab:cxr}).
The agent subsequently verified that this finding constitutes an independent diagnostic signal, completely distinct from broadly brighter or generally textured lungs.
When evaluated in a multivariable model, the spatial metric provides significant synergistic gains over mean entropy alone, yielding a peak area under the curve (AUC) of $0.851$ on the held-out set.
Furthermore, this structured interpretation decisively outperforms naive raw-pixel baselines, confirming that the predictive power stems directly from the system's high-level spatial abstraction.
The finding also proves highly resilient to hyperparameter variations, maintaining strong statistical significance across ablations of the sliding-window scale and bounding-box dimensions, successfully satisfying rigorous Bonferroni correction throughout all multiple-hypothesis testing.

The formulation of this hypothesis provides compelling evidence of genuine perceptual capability.
Identifying spatial heterogeneity demands direct visual engagement with the radiograph's intrinsic properties.
The system must actively observe these spatial patterns to conceptualize the metric prior to any mathematical quantification, a critical inductive step entirely inaccessible to text-bound statistical data mining.

\begin{table}[!t]
  \centering
  \caption{Numbers the system produced for the paediatric chest radiograph
    study. As in \cref{tab:stead}, every value is traceable to a real
    \texttt{run\_python} standard output.}
  \label{tab:cxr}
  \begin{tabular}{@{}lc@{\hspace{1.8em}}lc@{}}
    \toprule
    \multicolumn{2}{@{}l}{\textbf{\textcolor{tblBlue}{Effect and generalisation}}} &
    \multicolumn{2}{l@{}}{\textbf{\textcolor{tblBlue}{Discrimination and robustness}}} \\
    \cmidrule(r{1.8em}){1-2}\cmidrule{3-4}
    Patchiness effect size, all images  & $d=1.25$   & AUC, mean entropy only      & 0.840                 \\
    \quad development split             & $d=1.26$   & AUC, mean $+$ patchiness    & 0.851                 \\
    \quad held-out split                & $d=1.29$   & AUC, patchiness only        & 0.847                 \\
    Label difference, Mann-Whitney      & $p<0.0001$ & AUC, raw-pixel baseline     & 0.634                 \\
    Independent of the mean level       & $p=1.7\times10^{-5}$ & Window ablation, 8 / 16 / 32\,px & $d=1.63$ / $1.50$ / $1.36$ \\
                                        &            & Region-of-interest sweep    & \textcolor{nuswarm}{$d=1.18$ to $1.35$} \\
    \bottomrule
  \end{tabular}
\end{table}

\section{Conclusion}
\label{sec:conclusion}
In this paper, we present \OmniName, an end-to-end, omni-modal, and discipline-agnostic AI scientist. By integrating multimodal perception directly into the research lifecycle, the system enables raw observations to drive ideation, steer experimental execution, and substantiate the claims of the resulting manuscript. The framework demonstrates extensive cross-disciplinary applicability by successfully completing a 36-case demonstration suite spanning 5 discipline families. Comprehensive evaluations across 9 model backbones validate its capabilities, and paired ablations establish that direct multimodal perception is inherently decisive for scientific discovery. Ultimately, \OmniName establishes a foundational blueprint for future AI scientists and the broader development of automated empirical research.

\bibliographystyle{plainnat}
\bibliography{references}

\begin{thebibliography}{101}
\providecommand{\natexlab}[1]{#1}
\providecommand{\url}[1]{\texttt{#1}}
\expandafter\ifx\csname urlstyle\endcsname\relax
  \providecommand{\doi}[1]{doi: #1}\else
  \providecommand{\doi}{doi: \begingroup \urlstyle{rm}\Url}\fi

\bibitem[Alampara et~al.(2025)Alampara, Schilling-Wilhelmi, R{\'i}os-Garc{\'i}a, Mandal, Khetarpal, Grover, et~al.]{alampara2025macbench}
Nawaf Alampara, Mara Schilling-Wilhelmi, Marti{\~n}o R{\'i}os-Garc{\'i}a, Indrajeet Mandal, Pranav Khetarpal, Hargun~Singh Grover, et~al.
\newblock Probing the limitations of multimodal language models for chemistry and materials research.
\newblock \emph{Nature Computational Science}, 5:\penalty0 952--961, 2025.
\newblock \doi{10.1038/s43588-025-00836-3}.

\bibitem[Alayrac et~al.(2022)Alayrac, Donahue, Luc, Miech, Barr, Hasson, Lenc, Mensch, Millican, Reynolds, et~al.]{alayrac2022flamingo}
Jean-Baptiste Alayrac, Jeff Donahue, Pauline Luc, Antoine Miech, Iain Barr, Yana Hasson, Karel Lenc, Arthur Mensch, Katherine Millican, Malcolm Reynolds, et~al.
\newblock {Flamingo}: A visual language model for few-shot learning.
\newblock In \emph{Advances in Neural Information Processing Systems (NeurIPS)}, 2022.

\bibitem[{Anthropic}(2024)]{anthropic2024claude3}
{Anthropic}.
\newblock The {Claude} 3 model family: Opus, sonnet, haiku.
\newblock Anthropic, 2024.
\newblock Model card.

\bibitem[{Anthropic}(2026)]{claude}
{Anthropic}.
\newblock Claude.
\newblock Anthropic, 2026.
\newblock https://www.anthropic.com/claude.

\bibitem[Aversa et~al.(2018)Aversa, Modarres, Cozzini, Ciancio, and Chiusole]{data_nffa_sem}
Rossella Aversa, Mohammad~Hadi Modarres, Stefano Cozzini, Regina Ciancio, and Alberto Chiusole.
\newblock The first annotated set of scanning electron microscopy images for nanoscience.
\newblock \emph{Scientific Data}, 5:\penalty0 180172, 2018.
\newblock \doi{10.1038/sdata.2018.172}.

\bibitem[Baltru{\v{s}}aitis et~al.(2019)Baltru{\v{s}}aitis, Ahuja, and Morency]{baltrusaitis2019multimodal}
Tadas Baltru{\v{s}}aitis, Chaitanya Ahuja, and Louis-Philippe Morency.
\newblock Multimodal machine learning: A survey and taxonomy.
\newblock \emph{IEEE Transactions on Pattern Analysis and Machine Intelligence}, 41\penalty0 (2):\penalty0 423--443, 2019.
\newblock \doi{10.1109/TPAMI.2018.2798607}.

\bibitem[Baumgardner et~al.(2015)Baumgardner, Biehl, and Landgrebe]{data_indianpines}
Marion~F. Baumgardner, Larry~L. Biehl, and David~A. Landgrebe.
\newblock 220 band {AVIRIS} hyperspectral image data set: June 12, 1992 indian pine test site 3, 2015.

\bibitem[Behley et~al.(2019)Behley, Garbade, Milioto, Quenzel, Behnke, Stachniss, and Gall]{data_semantickitti}
Jens Behley, Martin Garbade, Andres Milioto, Jan Quenzel, Sven Behnke, Cyrill Stachniss, and J{\"u}rgen Gall.
\newblock {SemanticKITTI}: A dataset for semantic scene understanding of {LiDAR} sequences.
\newblock In \emph{IEEE/CVF International Conference on Computer Vision (ICCV)}, 2019.

\bibitem[Boiko et~al.(2023)Boiko, MacKnight, Kline, and Gomes]{boiko2023coscientist}
Daniil~A. Boiko, Robert MacKnight, Ben Kline, and Gabe Gomes.
\newblock Autonomous chemical research with large language models.
\newblock \emph{Nature}, 624\penalty0 (7992):\penalty0 570--578, 2023.

\bibitem[Burgess et~al.(2025)Burgess, Nirschl, Bravo-S{\'a}nchez, Lozano, Gupte, Galaz-Montoya, et~al.]{burgess2025microvqa}
James Burgess, Jeffrey~J. Nirschl, Laura Bravo-S{\'a}nchez, Alejandro Lozano, Sanket~Rajan Gupte, Jesus~G. Galaz-Montoya, et~al.
\newblock {MicroVQA}: A multimodal reasoning benchmark for microscopy-based scientific research.
\newblock In \emph{Proceedings of the IEEE/CVF Conference on Computer Vision and Pattern Recognition (CVPR)}, 2025.

\bibitem[Chan et~al.(2025)Chan, Chowdhury, Jaffe, Aung, Sherburn, Mays, et~al.]{chan2024mlebench}
Jun~Shern Chan, Neil Chowdhury, Oliver Jaffe, James Aung, Dane Sherburn, Evan Mays, et~al.
\newblock {MLE}-bench: Evaluating machine learning agents on machine learning engineering.
\newblock In \emph{International Conference on Learning Representations (ICLR)}, 2025.

\bibitem[Chen et~al.(2024)Chen, Ding, Lu, Williamson, Jaume, Song, et~al.]{chen2024uni}
Richard~J. Chen, Tong Ding, Ming~Y. Lu, Drew F.~K. Williamson, Guillaume Jaume, Andrew~H. Song, et~al.
\newblock Towards a general-purpose foundation model for computational pathology.
\newblock \emph{Nature Medicine}, 30:\penalty0 850--862, 2024.
\newblock \doi{10.1038/s41591-024-02857-3}.

\bibitem[Chen et~al.(2025)Chen, Chen, Ning, Zhang, Wang, Yu, et~al.]{chen2024scienceagentbench}
Ziru Chen, Shijie Chen, Yuting Ning, Qianheng Zhang, Boshi Wang, Botao Yu, et~al.
\newblock {ScienceAgentBench}: Toward rigorous assessment of language agents for data-driven scientific discovery.
\newblock In \emph{International Conference on Learning Representations (ICLR)}, 2025.

\bibitem[Darvish et~al.(2025)Darvish, Skreta, Zhao, Yoshikawa, Som, Bogdanovi{\'c}, et~al.]{darvish2024organa}
Kourosh Darvish, Marta Skreta, Yuchi Zhao, Naruki Yoshikawa, Sagnik Som, Miroslav Bogdanovi{\'c}, et~al.
\newblock {ORGANA}: A robotic assistant for automated chemistry experimentation and characterization.
\newblock \emph{Matter}, 8:\penalty0 101897, 2025.
\newblock \doi{10.1016/j.matt.2024.10.015}.

\bibitem[{DeepSeek-AI}(2025)]{deepseekr1}
{DeepSeek-AI}.
\newblock {DeepSeek-R1} incentivizes reasoning in {LLM}s through reinforcement learning.
\newblock \emph{Nature}, 645:\penalty0 633--638, 2025.
\newblock \doi{10.1038/s41586-025-09422-z}.

\bibitem[Flack et~al.(2016)Flack, Fiedler, Blas, et~al.]{data_whitestork}
Andrea Flack, Wolfgang Fiedler, Julio Blas, et~al.
\newblock Costs of migratory decisions: A comparison across eight white stork populations.
\newblock \emph{Science Advances}, 2\penalty0 (1):\penalty0 e1500931, 2016.
\newblock \doi{10.1126/sciadv.1500931}.

\bibitem[Gandhi et~al.(2025)Gandhi, Bolliet, and Zubeldia]{gandhi2025cmbagentvlm}
Kahaan Gandhi, Boris Bolliet, and Inigo Zubeldia.
\newblock Enhancing agentic autonomous scientific discovery with vision-language model capabilities.
\newblock \emph{arXiv preprint arXiv:2511.14631}, 2025.

\bibitem[Ghareeb et~al.(2026)Ghareeb, Chang, Mitchener, Yiu, Szostkiewicz, Shved, Gyimesi, Laurent, et~al.]{ghareeb2026robin}
Ali~E. Ghareeb, Benjamin Chang, Ludovico Mitchener, Angela Yiu, Caralyn~J. Szostkiewicz, Dmytro Shved, Gavin~J. Gyimesi, Jon~M. Laurent, et~al.
\newblock A multi-agent system for automating scientific discovery.
\newblock \emph{Nature}, 655:\penalty0 497--505, 2026.
\newblock \doi{10.1038/s41586-026-10652-y}.

\bibitem[{Google}(2026)]{gemma4}
{Google}.
\newblock Gemma 4 technical report, 2026.

\bibitem[Gottweis et~al.(2026)Gottweis, Weng, Daryin, Tu, Sirkovic, Myaskovsky, et~al.]{gottweis2026coscientist}
Juraj Gottweis, Wei-Hung Weng, Alexander Daryin, Tao Tu, Petar Sirkovic, Artiom Myaskovsky, et~al.
\newblock Accelerating scientific discovery with {Co-Scientist}.
\newblock \emph{Nature}, 655:\penalty0 487--496, 2026.
\newblock \doi{10.1038/s41586-026-10644-y}.

\bibitem[Hamidieh(2018)]{data_supercon}
Kam Hamidieh.
\newblock A data-driven statistical model for predicting the critical temperature of a superconductor.
\newblock \emph{Computational Materials Science}, 154:\penalty0 346--354, 2018.
\newblock \doi{10.1016/j.commatsci.2018.07.052}.

\bibitem[Helber et~al.(2019)Helber, Bischke, Dengel, and Borth]{data_eurosat}
Patrick Helber, Benjamin Bischke, Andreas Dengel, and Damian Borth.
\newblock {EuroSAT}: A novel dataset and deep learning benchmark for land use and land cover classification.
\newblock \emph{IEEE Journal of Selected Topics in Applied Earth Observations and Remote Sensing}, 12\penalty0 (7):\penalty0 2217--2226, 2019.
\newblock \doi{10.1109/JSTARS.2019.2918242}.

\bibitem[Hendricks et~al.(2020)Hendricks, Tkaczyk, Lin, and Feeney]{hendricks2020crossref}
Ginny Hendricks, Dominika Tkaczyk, Jennifer Lin, and Patricia Feeney.
\newblock {Crossref}: The sustainable source of community-owned scholarly metadata.
\newblock \emph{Quantitative Science Studies}, 1\penalty0 (1):\penalty0 414--427, 2020.

\bibitem[Hu et~al.(2020)Hu, Fey, Zitnik, Dong, Ren, Liu, Catasta, and Leskovec]{data_ogb}
Weihua Hu, Matthias Fey, Marinka Zitnik, Yuxiao Dong, Hongyu Ren, Bowen Liu, Michele Catasta, and Jure Leskovec.
\newblock Open graph benchmark: Datasets for machine learning on graphs.
\newblock In \emph{Advances in Neural Information Processing Systems (NeurIPS)}, volume~33, 2020.

\bibitem[Hughes and Salath{\'e}(2015)]{data_plantvillage}
David~P. Hughes and Marcel Salath{\'e}.
\newblock An open access repository of images on plant health to enable the development of mobile disease diagnostics through machine learning and crowdsourcing.
\newblock \emph{arXiv preprint arXiv:1511.08060}, 2015.

\bibitem[{InternAgent Team} et~al.(2025){InternAgent Team}, Zhang, Feng, Yan, Yuan, Ma, et~al.]{internagent2025}
{InternAgent Team}, Bo~Zhang, Shiyang Feng, Xiangchao Yan, Jiakang Yuan, Runmin Ma, et~al.
\newblock {InternAgent}: When agent becomes the scientist. building closed-loop system from hypothesis to verification.
\newblock \emph{arXiv preprint arXiv:2505.16938}, 2025.

\bibitem[{Intology AI}(2025)]{intology2025zochi}
{Intology AI}.
\newblock Zochi technical report, 2025.
\newblock Technical report, Intology AI.

\bibitem[Jakubik et~al.(2023)]{jakubik2023prithvi}
Johannes Jakubik et~al.
\newblock Foundation models for generalist geospatial artificial intelligence.
\newblock \emph{arXiv:2310.18660}, 2023.

\bibitem[Jiang et~al.(2025)Jiang, Schmidt, Srikanth, Xu, Kaplan, Jacenko, et~al.]{jiang2025aide}
Zhengyao Jiang, Dominik Schmidt, Dhruv Srikanth, Dixing Xu, Ian Kaplan, Deniss Jacenko, et~al.
\newblock {AIDE}: {AI}-driven exploration in the space of code.
\newblock \emph{arXiv preprint arXiv:2502.13138}, 2025.

\bibitem[Jumper et~al.(2021)Jumper, Evans, Pritzel, Green, Figurnov, Ronneberger, Tunyasuvunakool, Bates, et~al.]{jumper2021alphafold}
John Jumper, Richard Evans, Alexander Pritzel, Tim Green, Michael Figurnov, Olaf Ronneberger, Kathryn Tunyasuvunakool, Russ Bates, et~al.
\newblock Highly accurate protein structure prediction with {AlphaFold}.
\newblock \emph{Nature}, 596\penalty0 (7873):\penalty0 583--589, 2021.

\bibitem[Kather et~al.(2016)Kather, Weis, Bianconi, et~al.]{data_kather}
Jakob~Nikolas Kather, Cleo-Aron Weis, Francesco Bianconi, et~al.
\newblock Multi-class texture analysis in colorectal cancer histology.
\newblock \emph{Scientific Reports}, 6:\penalty0 27988, 2016.
\newblock \doi{10.1038/srep27988}.

\bibitem[Kay et~al.(2022)Kay, Kulits, Stathatos, et~al.]{data_cfc}
Justin Kay, Peter Kulits, Suzanne Stathatos, et~al.
\newblock The {Caltech Fish Counting} dataset: A benchmark for multiple-object tracking and counting.
\newblock In \emph{European Conference on Computer Vision (ECCV)}, 2022.

\bibitem[Kemp et~al.(2000)Kemp, Zwinderman, Tuk, Kamphuisen, and Ober{\'e}y{\'e}]{data_sleepedf}
Bob Kemp, Aeilko~H. Zwinderman, Bert Tuk, Hilbert A.~C. Kamphuisen, and Josefien J.~L. Ober{\'e}y{\'e}.
\newblock Analysis of a sleep-dependent neuronal feedback loop: the slow-wave microcontinuity of the {EEG}.
\newblock \emph{IEEE Transactions on Biomedical Engineering}, 47\penalty0 (9):\penalty0 1185--1194, 2000.
\newblock \doi{10.1109/10.867928}.

\bibitem[Kermany et~al.(2018)Kermany, Goldbaum, Cai, et~al.]{data_kermany}
Daniel~S. Kermany, Michael Goldbaum, Wenjia Cai, et~al.
\newblock Identifying medical diagnoses and treatable diseases by image-based deep learning.
\newblock \emph{Cell}, 172\penalty0 (5):\penalty0 1122--1131.e9, 2018.
\newblock \doi{10.1016/j.cell.2018.02.010}.

\bibitem[Kim et~al.(2020)Kim, Chi, Hu, Huang, and Ramani]{data_mcb}
Sangpil Kim, Hyung-gun Chi, Xiao Hu, Qixing Huang, and Karthik Ramani.
\newblock A large-scale annotated mechanical components benchmark for classification and retrieval tasks with deep neural networks.
\newblock In \emph{European Conference on Computer Vision (ECCV)}, pages 175--191, 2020.
\newblock \doi{10.1007/978-3-030-58523-5\_11}.

\bibitem[Kim et~al.(2025)Kim, Chen, Cheng, Gindulyte, He, He, Li, Shoemaker, Thiessen, Yu, et~al.]{data_pubchem}
Sunghwan Kim, Jie Chen, Tiejun Cheng, Asta Gindulyte, Jia He, Siqian He, Qingliang Li, Benjamin~A. Shoemaker, Paul~A. Thiessen, Bo~Yu, et~al.
\newblock {PubChem} 2025 update.
\newblock \emph{Nucleic Acids Research}, 53\penalty0 (D1):\penalty0 D1516--D1525, 2025.
\newblock \doi{10.1093/nar/gkae1059}.

\bibitem[{Kimi}(2025)]{kimik2}
{Kimi}.
\newblock {Kimi K2}: Open agentic intelligence, 2025.

\bibitem[Knapp et~al.(2010)Knapp, Kruk, Levinson, Diamond, and Neumann]{data_ibtracs}
Kenneth~R. Knapp, Michael~C. Kruk, David~H. Levinson, Howard~J. Diamond, and Charles~J. Neumann.
\newblock The international best track archive for climate stewardship ({IBTrACS}): Unifying tropical cyclone data.
\newblock \emph{Bulletin of the American Meteorological Society}, 91\penalty0 (3):\penalty0 363--376, 2010.
\newblock \doi{10.1175/2009BAMS2755.1}.

\bibitem[Lafuente et~al.(2015)Lafuente, Downs, Yang, and Stone]{data_rruff}
Barbara Lafuente, Robert~T. Downs, Hexiong Yang, and Nathan Stone.
\newblock The power of databases: the {RRUFF} project.
\newblock In Thomas Armbruster and Rosa~Micaela Danisi, editors, \emph{Highlights in Mineralogical Crystallography}, pages 1--30. W. De Gruyter, Berlin, Germany, 2015.
\newblock \doi{10.1515/9783110417104-003}.

\bibitem[Li et~al.(2026{\natexlab{a}})Li, Wu, Ji, Zhang, Fei, Zhang, Lee, and Hsu]{li2026taming}
Bobo Li, Rui Wu, Zibo Ji, Meishan Zhang, Hao Fei, Min Zhang, Mong-Li Lee, and Wynne Hsu.
\newblock Taming actor-observer asymmetry in agents via dialectical alignment.
\newblock In \emph{Proceedings of ACL}, pages 24068--24084, 2026{\natexlab{a}}.

\bibitem[Li and Zhao(2026)]{li2026auto}
Shawn Li and Yue Zhao.
\newblock The autonomy tax: Defense training breaks llm agents, 2026.
\newblock URL \url{https://arxiv.org/abs/2603.19423}.

\bibitem[Li et~al.(2026{\natexlab{b}})Li, Yu, Wang, Yang, Rossi, Dernoncourt, Hu, Yu, Xiao, Zhang, and Zhao]{li2026fortis}
Shawn Li, Chenxiao Yu, Han Wang, Wei Yang, Ryan Rossi, Franck Dernoncourt, Xiyang Hu, Philip Yu, Chaowei Xiao, Huan Zhang, and Yue Zhao.
\newblock Fortis: Benchmarking over-privilege in agent skills, 2026{\natexlab{b}}.
\newblock URL \url{https://arxiv.org/abs/2605.09163}.

\bibitem[Li et~al.(2024)Li, Yang, Choi, Zhu, Hsieh, Kim, et~al.]{li2024mmsci}
Zekun Li, Xianjun Yang, Kyuri Choi, Wanrong Zhu, Ryan Hsieh, HyeonJung Kim, et~al.
\newblock {MMSci}: A dataset for graduate-level multi-discipline multimodal scientific understanding.
\newblock \emph{arXiv preprint arXiv:2407.04903}, 2024.

\bibitem[{LIGO-Virgo Collaboration}(2021)]{data_gwosc}
{LIGO-Virgo Collaboration}.
\newblock Open data from the first and second observing runs of {Advanced LIGO} and {Advanced Virgo}.
\newblock \emph{SoftwareX}, 13:\penalty0 100658, 2021.
\newblock \doi{10.1016/j.softx.2021.100658}.

\bibitem[Lintott et~al.(2008)]{data_galaxyzoo}
Chris~J. Lintott et~al.
\newblock Galaxy zoo: morphologies derived from visual inspection of galaxies from the {Sloan Digital Sky Survey}.
\newblock \emph{Monthly Notices of the Royal Astronomical Society}, 389\penalty0 (3):\penalty0 1179--1189, 2008.
\newblock \doi{10.1111/j.1365-2966.2008.13689.x}.

\bibitem[Liu et~al.(2016)Liu, Springer, Li, Moody, et~al.]{data_heartsound}
Chengyu Liu, David Springer, Qiao Li, Benjamin Moody, et~al.
\newblock An open access database for the evaluation of heart sound algorithms.
\newblock \emph{Physiological Measurement}, 37\penalty0 (12):\penalty0 2181--2213, 2016.
\newblock \doi{10.1088/0967-3334/37/12/2181}.

\bibitem[Liu et~al.(2023)Liu, Li, Wu, and Lee]{liu2023llava}
Haotian Liu, Chunyuan Li, Qingyang Wu, and Yong~Jae Lee.
\newblock Visual instruction tuning.
\newblock In \emph{Advances in Neural Information Processing Systems (NeurIPS)}, 2023.

\bibitem[Lozano et~al.(2024)Lozano, Nirschl, Burgess, Gupte, Zhang, Unell, et~al.]{lozano2024mubench}
Alejandro Lozano, Jeffrey Nirschl, James Burgess, Sanket~Rajan Gupte, Yuhui Zhang, Alyssa Unell, et~al.
\newblock {$\mu$}-bench: A vision-language benchmark for microscopy understanding.
\newblock In \emph{Proceedings of NeurIPS (Datasets and Benchmarks Track)}, 2024.

\bibitem[Lu et~al.(2026)Lu, Lu, Lange, Yamada, Hu, Foerster, Ha, and Clune]{lu2026aiscientist}
Chris Lu, Cong Lu, Robert~Tjarko Lange, Yutaro Yamada, Shengran Hu, Jakob Foerster, David Ha, and Jeff Clune.
\newblock Towards end-to-end automation of {AI} research.
\newblock \emph{Nature}, 651:\penalty0 914--919, 2026.
\newblock \doi{10.1038/s41586-026-10265-5}.

\bibitem[Mandal et~al.(2025)Mandal, Soni, Zaki, Smedskjaer, Wondraczek, et~al.]{mandal2025aila}
Indrajeet Mandal, Jitendra Soni, Mohd Zaki, Morten~M. Smedskjaer, Katrin Wondraczek, et~al.
\newblock Evaluating large language model agents for automation of atomic force microscopy.
\newblock \emph{Nature Communications}, 16:\penalty0 9104, 2025.
\newblock \doi{10.1038/s41467-025-64105-7}.

\bibitem[Mitchener et~al.(2025)Mitchener, Yiu, Chang, Bourdenx, Nadolski, Sulovari, et~al.]{mitchener2025kosmos}
Ludovico Mitchener, Angela Yiu, Benjamin Chang, Mathieu Bourdenx, Tyler Nadolski, Arvis Sulovari, et~al.
\newblock {Kosmos}: An {AI} scientist for autonomous discovery.
\newblock \emph{arXiv preprint arXiv:2511.02824}, 2025.

\bibitem[Mousavi et~al.(2019)Mousavi, Sheng, Zhu, and Beroza]{data_stead}
S.~Mostafa Mousavi, Yixiao Sheng, Weiqiang Zhu, and Gregory~C. Beroza.
\newblock {STanford EArthquake Dataset (STEAD)}: A global data set of seismic signals for {AI}.
\newblock \emph{IEEE Access}, 7:\penalty0 179464--179476, 2019.
\newblock \doi{10.1109/ACCESS.2019.2947848}.

\bibitem[Nguyen et~al.(2016)Nguyen, Tran, Ngo, Phan, Lumbanraja, Faisal, Abapihi, Kubo, and Satou]{data_dna_histone}
Ngoc~Giang Nguyen, Vu~Anh Tran, Duc~Luu Ngo, Dau Phan, Favorisen~Rosyking Lumbanraja, Mohammad~Reza Faisal, Bahriddin Abapihi, Mamoru Kubo, and Kenji Satou.
\newblock {DNA} sequence classification by convolutional neural network.
\newblock \emph{Journal of Biomedical Science and Engineering}, 9\penalty0 (5):\penalty0 280--286, 2016.
\newblock \doi{10.4236/jbise.2016.95021}.

\bibitem[{OpenAI}(2023)]{openai2023gpt4}
{OpenAI}.
\newblock {GPT-4} technical report.
\newblock \emph{arXiv preprint arXiv:2303.08774}, 2023.

\bibitem[{OpenAI}(2026)]{gpt56}
{OpenAI}.
\newblock {GPT-5.6}.
\newblock OpenAI, 2026.

\bibitem[Orenstein et~al.(2015)Orenstein, Beijbom, Peacock, and Sosik]{data_whoiplankton}
Eric~C. Orenstein, Oscar Beijbom, Emily~E. Peacock, and Heidi~M. Sosik.
\newblock {WHOI-Plankton}: A large scale fine grained visual recognition benchmark dataset for plankton classification.
\newblock \emph{arXiv preprint arXiv:1510.00745}, 2015.

\bibitem[Parker et~al.(2024)]{parker2023astroclip}
Liam Parker et~al.
\newblock {AstroCLIP}: A cross-modal foundation model for galaxies.
\newblock \emph{Monthly Notices of the Royal Astronomical Society}, 531:\penalty0 4990--5011, 2024.
\newblock \doi{10.1093/mnras/stae1450}.

\bibitem[Pramanick et~al.(2024)Pramanick, Chellappa, and Venugopalan]{pramanick2024spiqa}
Shraman Pramanick, Rama Chellappa, and Subhashini Venugopalan.
\newblock {SPIQA}: A dataset for multimodal question answering on scientific papers.
\newblock In \emph{Advances in Neural Information Processing Systems (NeurIPS) Datasets and Benchmarks Track}, 2024.

\bibitem[Priem et~al.(2022)Priem, Piwowar, and Orr]{priem2022openalex}
Jason Priem, Heather Piwowar, and Richard Orr.
\newblock {OpenAlex}: A fully-open index of scholarly works, authors, venues, institutions, and concepts.
\newblock \emph{arXiv preprint arXiv:2205.01833}, 2022.

\bibitem[Prodanovi{\'c} et~al.(2015)Prodanovi{\'c}, Esteva, Hanlon, Nanda, and Agarwal]{data_digitalrocks}
Ma{\v s}a Prodanovi{\'c}, Maria Esteva, Matthew Hanlon, Ganesh Nanda, and Prateek Agarwal.
\newblock Digital rocks portal: a repository for porous media images.
\newblock Digital Rocks Portal, University of Texas at Austin, 2015.

\bibitem[Purohit et~al.(2019)Purohit, Tanabe, Ichige, Endo, Nikaido, Suefusa, and Kawaguchi]{data_mimii}
Harsh Purohit, Ryo Tanabe, Kenji Ichige, Takashi Endo, Yuki Nikaido, Kaori Suefusa, and Yohei Kawaguchi.
\newblock {MIMII} dataset: Sound dataset for malfunctioning industrial machine investigation and inspection.
\newblock In \emph{Workshop on Detection and Classification of Acoustic Scenes and Events (DCASE)}, 2019.

\bibitem[{Qwen Team}(2026)]{qwen3}
{Qwen Team}.
\newblock {Qwen3.5}.
\newblock Alibaba Qwen, 2026.
\newblock \url{https://huggingface.co/Qwen}.

\bibitem[Radford et~al.(2021)Radford, Kim, Hallacy, Ramesh, Goh, Agarwal, Sastry, Askell, Mishkin, Clark, Krueger, and Sutskever]{radford2021clip}
Alec Radford, Jong~Wook Kim, Chris Hallacy, Aditya Ramesh, Gabriel Goh, Sandhini Agarwal, Girish Sastry, Amanda Askell, Pamela Mishkin, Jack Clark, Gretchen Krueger, and Ilya Sutskever.
\newblock Learning transferable visual models from natural language supervision.
\newblock In \emph{International Conference on Machine Learning (ICML)}, 2021.

\bibitem[Roberts et~al.(2024)Roberts, Han, Houlsby, and Albanie]{roberts2024scifibench}
Jonathan Roberts, Kai Han, Neil Houlsby, and Samuel Albanie.
\newblock {SciFIBench}: Benchmarking large multimodal models for scientific figure interpretation.
\newblock In \emph{Proceedings of NeurIPS (Datasets and Benchmarks Track)}, 2024.

\bibitem[Robinson et~al.(2025)Robinson, Miron, Hagiwara, Weck, Keen, Alizadeh, et~al.]{robinson2024naturelm}
David Robinson, Marius Miron, Masato Hagiwara, Benno Weck, Sara Keen, Milad Alizadeh, et~al.
\newblock {NatureLM}-audio: An audio-language foundation model for bioacoustics.
\newblock In \emph{International Conference on Learning Representations (ICLR)}, 2025.

\bibitem[Sayigh et~al.(2016)Sayigh, Daher, Allen, Gordon, Joyce, Stuhlmann, and Tyack]{data_watkins}
Laela Sayigh, Mary~Ann Daher, Julie Allen, Helen Gordon, Katherine Joyce, Claire Stuhlmann, and Peter Tyack.
\newblock The {Watkins Marine Mammal Sound Database}: An online, freely accessible resource.
\newblock \emph{Proceedings of Meetings on Acoustics}, 27\penalty0 (1):\penalty0 040013, 2016.
\newblock \doi{10.1121/2.0000358}.

\bibitem[Sch{\"a}fer et~al.(2018)Sch{\"a}fer, Santana, Haden, and Biasini]{data_comma2k19}
Harald Sch{\"a}fer, Eder Santana, Andrew Haden, and Riccardo Biasini.
\newblock A commute in data: The comma2k19 dataset.
\newblock \emph{arXiv preprint arXiv:1812.05752}, 2018.

\bibitem[Schick et~al.(2023)Schick, Dwivedi-Yu, Dess{\`i}, Raileanu, Lomeli, Hambro, Zettlemoyer, Cancedda, and Scialom]{schick2023toolformer}
Timo Schick, Jane Dwivedi-Yu, Roberto Dess{\`i}, Roberta Raileanu, Maria Lomeli, Eric Hambro, Luke Zettlemoyer, Nicola Cancedda, and Thomas Scialom.
\newblock {Toolformer}: Language models can teach themselves to use tools.
\newblock In \emph{Advances in Neural Information Processing Systems (NeurIPS)}, 2023.

\bibitem[Schmidgall et~al.(2025)]{schmidgall2025agentlab}
Samuel Schmidgall et~al.
\newblock Agent laboratory: Using {LLM} agents as research assistants.
\newblock \emph{arXiv preprint arXiv:2501.04227}, 2025.

\bibitem[Schunck et~al.(2021)Schunck, Magistri, et~al.]{data_pheno4d}
David Schunck, Federico Magistri, et~al.
\newblock {Pheno4D}: A spatio-temporal dataset of maize and tomato plant point clouds for phenotyping and advanced plant analysis.
\newblock \emph{PLOS ONE}, 16\penalty0 (8):\penalty0 e0256340, 2021.
\newblock \doi{10.1371/journal.pone.0256340}.

\bibitem[Shao et~al.(2025)Shao, Huang, Li, Zhao, Xu, Li, Liu, et~al.]{shao2025omniscientist}
Chenyang Shao, Dehao Huang, Yu~Li, Keyu Zhao, Fengli Xu, Yong Li, Tie-Yan Liu, et~al.
\newblock {OmniScientist}: Toward a co-evolving ecosystem of human and {AI} scientists.
\newblock \emph{arXiv preprint arXiv:2511.16931}, 2025.

\bibitem[Shinn et~al.(2023)Shinn, Cassano, Berman, Gopinath, Narasimhan, and Yao]{shinn2023reflexion}
Noah Shinn, Federico Cassano, Edward Berman, Ashwin Gopinath, Karthik Narasimhan, and Shunyu Yao.
\newblock {Reflexion}: Language agents with verbal reinforcement learning.
\newblock In \emph{Advances in Neural Information Processing Systems (NeurIPS)}, 2023.

\bibitem[Starace et~al.(2025)Starace, Jaffe, Sherburn, Aung, Chan, Maksin, et~al.]{starace2025paperbench}
Giulio Starace, Oliver Jaffe, Dane Sherburn, James Aung, Jun~Shern Chan, Leon Maksin, et~al.
\newblock {PaperBench}: Evaluating {AI}'s ability to replicate {AI} research.
\newblock \emph{arXiv preprint arXiv:2504.01848}, 2025.

\bibitem[Stowell et~al.(2019)Stowell, Wood, Pamu{\l}a, Stylianou, and Glotin]{data_birdaudio}
Dan Stowell, Michael~D. Wood, Hanna Pamu{\l}a, Yannis Stylianou, and Herv{\'e} Glotin.
\newblock Automatic acoustic detection of birds through deep learning: the first {Bird Audio Detection} challenge.
\newblock \emph{Methods in Ecology and Evolution}, 10\penalty0 (3):\penalty0 368--380, 2019.
\newblock \doi{10.1111/2041-210X.13103}.

\bibitem[Sun et~al.(2021)Sun, Karigo, Chakraborty, et~al.]{data_calms21}
Jennifer~J. Sun, Tomomi Karigo, Dipam Chakraborty, et~al.
\newblock The multi-agent behavior dataset: Mouse dyadic social interactions.
\newblock In \emph{Advances in Neural Information Processing Systems (NeurIPS) Datasets and Benchmarks Track}, 2021.

\bibitem[Sun et~al.(2026)Sun, Liu, Ma, et~al.]{sun2025scienceboard}
Qiushi Sun, Zhoumianze Liu, Chang Ma, et~al.
\newblock {ScienceBoard}: Evaluating multimodal autonomous agents in realistic scientific workflows.
\newblock In \emph{International Conference on Learning Representations (ICLR)}, 2026.

\bibitem[Swanson et~al.(2025)Swanson, Wu, Bulaong, Pak, and Zou]{swanson2025virtuallab}
Kyle Swanson, Wesley Wu, Nash~L. Bulaong, John~E. Pak, and James Zou.
\newblock The virtual lab of {AI} agents designs new {SARS}-{CoV}-2 nanobodies.
\newblock \emph{Nature}, 646:\penalty0 716--723, 2025.
\newblock \doi{10.1038/s41586-025-09442-9}.

\bibitem[Szymanski et~al.(2023)Szymanski, Rendy, Fei, Kumar, He, Milsted, et~al.]{szymanski2023alab}
Nathan~J. Szymanski, Bernardus Rendy, Yuxing Fei, Rishi~E. Kumar, Tanjin He, David Milsted, et~al.
\newblock An autonomous laboratory for the accelerated synthesis of inorganic materials.
\newblock \emph{Nature}, 624:\penalty0 86--91, 2023.
\newblock \doi{10.1038/s41586-023-06734-w}.

\bibitem[Takamoto et~al.(2022)Takamoto, Praditia, Leiteritz, MacKinlay, Alesiani, Pfl{\"u}ger, and Niepert]{data_pdebench}
Makoto Takamoto, Timothy Praditia, Raphael Leiteritz, Dan MacKinlay, Francesco Alesiani, Dirk Pfl{\"u}ger, and Mathias Niepert.
\newblock {PDEBench}: An extensive benchmark for scientific machine learning.
\newblock In \emph{Advances in Neural Information Processing Systems (NeurIPS) Datasets and Benchmarks Track}, 2022.

\bibitem[Tang et~al.(2025)Tang, Xia, Li, and Huang]{tang2025airesearcher}
Jiabin Tang, Lianghao Xia, Zhonghang Li, and Chao Huang.
\newblock {AI-Researcher}: Autonomous scientific innovation.
\newblock \emph{arXiv preprint arXiv:2505.18705}, 2025.

\bibitem[Udrescu and Tegmark(2020)]{data_feynman}
Silviu-Marian Udrescu and Max Tegmark.
\newblock {AI Feynman}: A physics-inspired method for symbolic regression.
\newblock \emph{Science Advances}, 6\penalty0 (16):\penalty0 eaay2631, 2020.
\newblock \doi{10.1126/sciadv.aay2631}.

\bibitem[Ulman et~al.(2017)Ulman, Ma{\v s}ka, Magnusson, et~al.]{data_celltrack}
Vladim{\'i}r Ulman, Martin Ma{\v s}ka, Klas E.~G. Magnusson, et~al.
\newblock An objective comparison of cell-tracking algorithms.
\newblock \emph{Nature Methods}, 14:\penalty0 1141--1152, 2017.
\newblock \doi{10.1038/nmeth.4473}.

\bibitem[{U.S. DOT FHWA}(2016)]{data_ngsim}
{U.S. DOT FHWA}.
\newblock Next generation simulation ({NGSIM}) vehicle trajectories and supporting data.
\newblock ITS DataHub (data.transportation.gov), 2016.

\bibitem[Veillette et~al.(2020)Veillette, Samsi, and Mattioli]{data_sevir}
Mark~S. Veillette, Siddharth Samsi, and Christopher~J. Mattioli.
\newblock {SEVIR}: A storm event imagery dataset for deep learning applications in radar and satellite meteorology.
\newblock In \emph{Proceedings of NeurIPS}, volume~33, 2020.

\bibitem[Villaescusa-Navarro et~al.(2025)Villaescusa-Navarro, Bolliet, Villanueva-Domingo, Bayer, Acquah, Amancharla, et~al.]{villaescusa2025denario}
Francisco Villaescusa-Navarro, Boris Bolliet, Pablo Villanueva-Domingo, Adrian~E. Bayer, Aidan Acquah, Chetana Amancharla, et~al.
\newblock The {Denario} project: Deep knowledge {AI} agents for scientific discovery.
\newblock \emph{arXiv:2510.26887}, 2025.

\bibitem[Walmsley et~al.(2022)]{data_gzdecals}
Mike Walmsley et~al.
\newblock Galaxy zoo {DECaLS}: Detailed visual morphology measurements from volunteers and deep learning for 314{,}000 galaxies.
\newblock \emph{Monthly Notices of the Royal Astronomical Society}, 509\penalty0 (3):\penalty0 3966--3988, 2022.
\newblock \doi{10.1093/mnras/stab2093}.

\bibitem[Wang et~al.(2023)Wang, Fu, Du, Gao, Huang, Liu, Chandak, Liu, Van~Katwyk, Deac, et~al.]{wang2023scidiscovery}
Hanchen Wang, Tianfan Fu, Yuanqi Du, Wenhao Gao, Kexin Huang, Ziming Liu, Payal Chandak, Shengchao Liu, Peter Van~Katwyk, Andreea Deac, et~al.
\newblock Scientific discovery in the age of artificial intelligence.
\newblock \emph{Nature}, 620:\penalty0 47--60, 2023.
\newblock \doi{10.1038/s41586-023-06221-2}.

\bibitem[Wang et~al.(2024)Wang, Xia, He, Chen, Liu, Zhu, et~al.]{wang2024charxiv}
Zirui Wang, Mengzhou Xia, Luxi He, Howard Chen, Yitao Liu, Richard Zhu, et~al.
\newblock {CharXiv}: Charting gaps in realistic chart understanding in multimodal {LLMs}.
\newblock In \emph{Proceedings of NeurIPS (Datasets and Benchmarks Track)}, 2024.

\bibitem[Wei et~al.(2022)Wei, Wang, Schuurmans, Bosma, Ichter, Xia, Chi, Le, and Zhou]{wei2022cot}
Jason Wei, Xuezhi Wang, Dale Schuurmans, Maarten Bosma, Brian Ichter, Fei Xia, Ed~H. Chi, Quoc~V. Le, and Denny Zhou.
\newblock Chain-of-thought prompting elicits reasoning in large language models.
\newblock In \emph{Proceedings of NeurIPS}, 2022.

\bibitem[Wei et~al.(2025)Wei, Yang, Zhang, Chen, Zhuang, Gao, et~al.]{wei2025agenticscience}
Jiaqi Wei, Yue-Jin Yang, Xiang Zhang, Yuhan Chen, Xiang Zhuang, Zhangyang Gao, et~al.
\newblock From {AI} for science to agentic science: A survey on autonomous scientific discovery.
\newblock \emph{arXiv preprint arXiv:2508.14111}, 2025.

\bibitem[Xu et~al.(2025)Xu, Wu, Yu, Yan, Jiang, He, et~al.]{xu2025expvid}
Yicheng Xu, Yue Wu, Jiashuo Yu, Ziang Yan, Tianxiang Jiang, Yinan He, et~al.
\newblock {ExpVid}: A benchmark for experiment video understanding and reasoning.
\newblock \emph{arXiv preprint arXiv:2510.11606}, 2025.

\bibitem[Yamada et~al.(2025)Yamada, Lange, Lu, Hu, Lu, Foerster, Clune, and Ha]{yamada2025aiscientist2}
Yutaro Yamada, Robert~Tjarko Lange, Cong Lu, Shengran Hu, Chris Lu, Jakob Foerster, Jeff Clune, and David Ha.
\newblock The {AI} {S}cientist-v2: Workshop-level automated scientific discovery via agentic tree search.
\newblock \emph{arXiv preprint arXiv:2504.08066}, 2025.

\bibitem[Yang et~al.(2023{\natexlab{a}})Yang, Shi, Wei, Liu, Zhao, Ke, Pfister, and Ni]{data_medmnist}
Jiancheng Yang, Rui Shi, Donglai Wei, Zequan Liu, Lin Zhao, Bilian Ke, Hanspeter Pfister, and Bingbing Ni.
\newblock {MedMNIST v2}: A large-scale lightweight benchmark for 2d and 3d biomedical image classification.
\newblock \emph{Scientific Data}, 10:\penalty0 41, 2023{\natexlab{a}}.
\newblock \doi{10.1038/s41597-022-01721-8}.

\bibitem[Yang et~al.(2023{\natexlab{b}})]{yang2023mmreact}
Zhengyuan Yang et~al.
\newblock {MM-ReAct}: Prompting {ChatGPT} for multimodal reasoning and action.
\newblock \emph{arXiv:2303.11381}, 2023{\natexlab{b}}.

\bibitem[Yao et~al.(2025)Yao, Samantray, Ghosh, Roccapriore, Kovarik, Allec, and Ziatdinov]{yao2025scilink}
Lance Yao, Suman Samantray, Ayana Ghosh, Kevin~M. Roccapriore, Libor Kovarik, Sarah~I. Allec, and Maxim Ziatdinov.
\newblock Operationalizing serendipity: Multi-agent {AI} workflows for enhanced materials characterization with theory-in-the-loop.
\newblock \emph{arXiv preprint arXiv:2508.06569}, 2025.

\bibitem[Yao et~al.(2023)Yao, Zhao, Yu, Du, Shafran, Narasimhan, and Cao]{yao2023react}
Shunyu Yao, Jeffrey Zhao, Dian Yu, Nan Du, Izhak Shafran, Karthik Narasimhan, and Yuan Cao.
\newblock {ReAct}: Synergizing reasoning and acting in language models.
\newblock In \emph{International Conference on Learning Representations (ICLR)}, 2023.

\bibitem[Yuan et~al.(2025)Yuan, Yan, Shi, et~al.]{yuan2025dolphin}
Jiakang Yuan, Xiangchao Yan, Botian Shi, et~al.
\newblock {Dolphin}: Moving towards closed-loop auto-research through thinking, practice, and feedback.
\newblock \emph{arXiv preprint arXiv:2501.03916}, 2025.

\bibitem[Yue et~al.(2024)]{yue2023mmmu}
Xiang Yue et~al.
\newblock {MMMU}: A massive multi-discipline multimodal understanding and reasoning benchmark for expert {AGI}.
\newblock In \emph{Proceedings of the IEEE/CVF Conference on Computer Vision and Pattern Recognition (CVPR)}, 2024.

\bibitem[Zhao et~al.(2026)Zhao, Beery, and Mac~Aodha]{zhao2026discoper}
Bingchen Zhao, Sara Beery, and Oisin Mac~Aodha.
\newblock Autonomous scientific discovery via iterative meta-reflection.
\newblock \emph{arXiv preprint arXiv:2607.01131}, 2026.

\bibitem[{Zhipu}(2026)]{glm5}
{Zhipu}.
\newblock {GLM}-5: From vibe coding to agentic engineering, 2026.

\bibitem[Zhou et~al.(2025)Zhou, Wang, He, Shen, Xiao, Li, et~al.]{zhou2025sfe}
Yuhao Zhou, Yiheng Wang, Xuming He, Ao~Shen, Ruoyao Xiao, Zhiwei Li, et~al.
\newblock Scientists' first exam: Probing cognitive abilities of {MLLM} via perception, understanding, and reasoning.
\newblock \emph{arXiv preprint arXiv:2506.10521}, 2025.

\end{thebibliography}

\appendix
\section{Additional evaluation detail}
\label{app:eval}
\begin{figure}[t]
  \centering
  \includegraphics[width=0.6\linewidth]{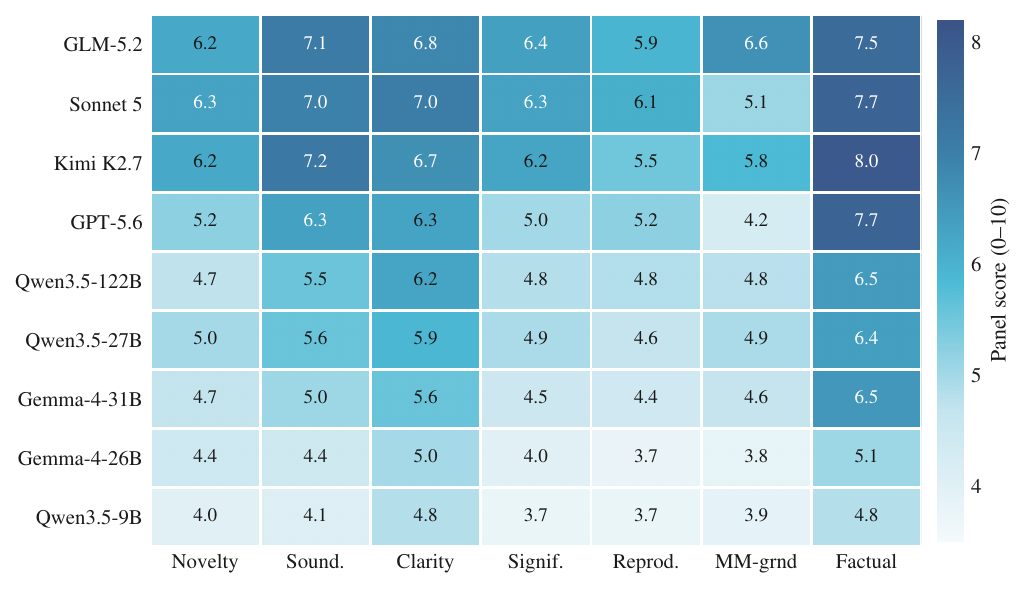}
  \caption{Heatmap visualization of the evaluation scores from \cref{tab:eval-backbone}. Darker cells indicate higher panel scores. Model reasoning strength corresponds to row darkness, with the strongest backbones positioned at the top and smaller open-weight models at the bottom. Notably, factual accuracy consistently achieves the highest scores (the darkest column) across all evaluated models.}
  \label{fig:eval-heatmap}
\end{figure}

This section provides the supporting analyses for the primary evaluation presented in \cref{sec:eval-main}.
These supplementary results include the per-paper computational cost across different backbones (\cref{tab:eval-cost}), a heatmap visualization of the backbone performance rubric (\cref{fig:eval-heatmap}), a dimension-by-dimension breakdown of the perception gain (\cref{tab:eval-perception-delta}), and the validation metrics for the judge panel (\cref{tab:eval-judge}).

\begin{table}[ht]
    \centering
    \caption{Cost per paper by backbone model. The experiment stage dominates the overall cost across all evaluated systems. $^\dagger$Open-weight backbones were run locally.}
    \label{tab:eval-cost}
    \begin{tabular}{@{}l >{\centering\arraybackslash}p{3.15cm} >{\centering\arraybackslash}p{2.5cm} >{\centering\arraybackslash}p{1.5cm} >{\centering\arraybackslash}p{2.65cm}@{}}
      \toprule
      \textbf{Backbone} & \textbf{Tokens in/out}\lob & \textbf{Cache hit}\hib & \textbf{\$ / paper}\lob & \textbf{Wall-clock}\lob \\
      \midrule
      Sonnet 5                 & 98k / 191k & 93\% & \$2.63 & 29 min \\
      GPT-5.6                  & 122k / 112k & 89\% & \$4.34 & 12 min \\
      Qwen3.5-27B$^{\dagger}$  & 1.4M / 75k & -- & \$0.06 & 30 min \\
      Gemma-4-31B$^{\dagger}$  & 714k / 41k & -- & \$0.03 & 14 min \\
      \bottomrule
    \end{tabular}
\end{table}

\begin{table}[ht]
  \centering \setlength{\tabcolsep}{4pt}\renewcommand{\arraystretch}{1.0}\setlength{\extrarowheight}{0pt}
  \caption{Perception gain per evaluation dimension. Across 5 blind pairs and 1 vision-off pair, each cell represents the difference in panel scores (original scale: $0$--$10$) between the perception-enabled system and the blind baseline. Positive values indicate that visual perception improves performance. $^\dagger$ Cardiology uses the weaker vision-off run.}
  \label{tab:eval-perception-delta}
  \begin{tabularx}{\linewidth}{@{}l *{5}{Y} YY Y@{}}
    \toprule
    & \multicolumn{5}{c}{\textbf{Standard peer-review}}
    & \multicolumn{2}{c}{\textbf{MM-mandatory}} & \\
    \cmidrule(lr){2-6}\cmidrule(lr){7-8}
    \textbf{Case ($\Delta$ = on $-$ baseline)} & Novelty\hib & Sound.\hib & Clarity\hib & Signif.\hib & Reprod.\hib & \mbox{MM-gr.\hib} & Factual\hib & \textbf{Overall}\hib \\
    \midrule
    Galaxy cross-survey & +2.5 & +1.2 & +1.0 & +2.3 & +0.7 & +1.3 & +2.0 & +1.7 \\
    \zb Seismology & +1.0 & +1.5 & -1.0 & +1.5 & +0.5 & +1.5 & +0.5 & +1.5 \\
    Pathology & +1.5 & +1.0 & +0.5 & +1.0 & +0.5 & +3.0 & -0.5 & +1.5 \\
    \zb Mechanical CAD & +0.5 & +0.5 & +3.5 & +2.5 & +2.5 & +1.0 & +2.0 & +3.0 \\
    Plant phenotyping & +1.0 & +0.0 & +0.0 & +0.5 & +0.7 & +4.0 & +0.3 & +0.3 \\
    \zb Cardiology$^{\dagger}$ & +0.3 & +0.3 & +0.3 & +0.0 & +1.3 & -1.7 & +0.0 & +0.3 \\
    \midrule
    \textbf{Macro-average $\Delta$} & \textbf{\textcolor{tblGain}{+1.14}} & \textbf{\textcolor{tblGain}{+0.75}} & \textbf{\textcolor{tblGain}{+0.72}} & \textbf{\textcolor{tblGain}{+1.31}} & \textbf{\textcolor{tblGain}{+1.03}} & \textbf{\textcolor{tblGain}{+1.53}} & \textbf{\textcolor{tblGain}{+0.72}} & \textbf{\textcolor{tblGain}{+1.39}} \\
    \bottomrule
  \end{tabularx}
\end{table}

\begin{table}[ht]
  \centering
  \caption{Validating the judge before it is trusted; the target column lists the
    pre-set acceptance thresholds.}
  \label{tab:eval-judge}
  \begin{tabular}{@{}llcc@{}}
    \toprule
    \textbf{Validity check} & \textbf{Statistic} & \textbf{Target} & \textbf{Measured} \\
    \midrule
    Inter-judge agreement     & Krippendorff $\alpha$ & $>0.6$      & 0.66 \\
    Self-preference bias      & own $-$ others        & $\approx 0$ & 0$^{\dagger}$ \\
    Verbosity bias            & score vs.\ length $\rho$ & $\approx 0$ & 0.16 \\
    \bottomrule
  \end{tabular}
\end{table}

\Cref{lst:judge} reproduces the rubric both judges receive. It is the whole of what they are told, besides the manuscript source, its figure captions, and the authors' result ledger, and the same text is used for every case and every backbone.

\prmtitle{Review rubric: the whole of what each judge is told}
\begin{lstlisting}[style=prompt,caption={The review rubric, verbatim. Each judge receives this, the manuscript source, the figure captions, and the authors' result ledger, and nothing else.},label={lst:judge}]
You are an expert, critical peer reviewer for a top venue, reviewing ONE paper produced by an
automated multimodal research system. Judge ONLY what is written in the paper and shown in its
figures. $Do NOT inflate scores.$

Score these @SEVEN dimensions@, each an INTEGER 1-10 (1=very poor, 10=excellent), and be strict --
an incremental or workshop-level paper must be scored as such, never rubber-stamped:
1. novelty -- originality against real prior art.
2. soundness -- method and statistics correct (controls, multiple-comparison correction, no
   leakage).
3. clarity -- presentation and structure.
4. significance -- importance of the finding.
5. reproducibility -- enough detail to re-run.
6. mm_grounding -- does the paper GENUINELY use the visual / observational evidence (the
   attached figures of raw observations), or is it just statistics on scalar features? Reward
   papers that SHOW and INTERPRET raw observations; penalize ones whose figures are absent,
   unreadable, or decorative.
7. factual_accuracy -- does EVERY headline number in the paper trace to the AUTHORS' RESULT
   LEDGER below? Penalize any statistic or claim not supported by the ledger.
\end{lstlisting}

\section{The checks enforced in code}
\label{app:checks}
The idea, rigour, and claim checks of \cref{sec:framework} are not model self-assessment.
Each is a Python predicate that receives the stage's \texttt{finalize} payload and the accumulated loop state, returning either acceptance or a reason string.
A rejection is appended to the conversation as a new observation and the agent continues; the stage cannot end until the predicate accepts or the step budget runs out.
\Cref{alg:stage-loop} details the loop, \cref{tab:gate-conditions} lists every condition, and \cref{tab:gate-fired} reports what those conditions rejected over the 36 primary runs.

\begin{algorithm}[!t]
\caption{One gated stage. The agent reasons freely inside the loop, while the stage boundary is a deterministic predicate over the run's own record. \Cref{alg:gate} expands the predicate $g$ for the experiment stage.}
\label{alg:stage-loop}
\begin{algorithmic}[1]
\Require system prompt $s$, task message $m$, tools $T$, exit predicate $g$, step budget $K$, perception budget $B$
\State $h \gets [s, m]$; \quad $\sigma \gets \{\textsc{good\_runs}{=}0,\ \textsc{searched}{=}0,\ \textsc{img\_used}{=}0,\ \textsc{stdout}{=}\varnothing\}$
\For{$k = 1$ \textbf{to} $K$}
  \State $a \gets \Call{Model}{h, T}$ \Comment{reasoning plus zero or more tool calls}
  \ForAll{tool calls $c \in a$}
    \If{$c$ is a perception call \textbf{and} $\sigma.\textsc{img\_used} \geq B$}
      \State $r \gets \text{``budget exhausted''}$
    \Else
      \State $r \gets \Call{Execute}{c}$; \quad update $\sigma$ \Comment{stdout, data reads, searches, looks}
    \EndIf
    \State append $r$ to $h$
  \EndFor
  \If{$a$ called \texttt{finalize}}
    \State $(\mathit{ok}, \mathit{why}) \gets g(a.\text{payload}, \sigma)$ \Comment{the check, in code, over what really happened}
    \If{$\mathit{ok}$}
      \State \Return $a.\text{payload}$ \Comment{stage output admitted}
    \Else
      \State append $\mathit{why}$ to $h$ \Comment{the agent is told what failed and continues}
    \EndIf
  \EndIf
\EndFor
\State \Return \textsc{Failed} \Comment{the outer pipeline backtracks to ideation}
\end{algorithmic}
\end{algorithm}

\begin{table}[p]
  \centering
  \setlength{\tabcolsep}{4pt}\renewcommand{\arraystretch}{1.06}
  \caption{Every condition enforced by the 3 checks. Each row is a separate predicate in the stage's exit function; failing any predicate returns the agent to the loop with the corresponding demand as the reason. The claim check runs on the drafted manuscript rather than a \texttt{finalize} payload, so it acts on the text itself.}
  \label{tab:gate-conditions}
  \begin{tabularx}{\linewidth}{@{}>{\raggedright\arraybackslash}p{2.6cm} X@{}}
    \toprule
    \textbf{Condition} & \textbf{What it demands of the stage output} \\
    \midrule
    \grpspanA{2}{Idea check (ideation)}
    Schema & A research question, a hypothesis, an experiment protocol, and a falsification criterion, all non-empty. \\
    \zb Breadth & At least 5 self-screened candidate projects, each rated for novelty risk. \\
    Prior art & At least 3 focused literature searches, one of them aimed at the selected idea specifically. \\
    \zb Feasibility & The selected idea marked fully computational. A proposal that would need a physical experiment is refused outright. \\
    Minimal claim & The smallest claim worth publishing if the rest of the study fails, stated separately from the hypothesis. \\
    \zb Novelty evidence & What the searches returned for and against this particular idea, with citations. \\
    Claim scope & An explicit statement of what the data cannot establish, separating the measured proxy from any mechanistic or causal reading. \\
    \zb Effective sample & The decisive-event count for the key test, estimated from the real data counts, and whether it is adequate. \\
    Leakage & Whether any step uses ground-truth labels at decision time, and what the label-agnostic counterpart is. \\
    \zb Visual audit & Required once the agent has looked at any raw item: which groups it viewed, and how a disagreement with the given label was resolved. \\
    Novelty language & Absolute-novelty phrasing (``first'', ``unstudied'', ``no prior work'') is rejected, because a bounded search cannot support it. \\
    \grpspan{2}{Rigour check (experiment)}
    Verdict & One of \emph{supported}, \emph{refuted}, \emph{mixed}, \emph{null}, or \emph{infeasible}. An honest negative is a valid exit. \\
    \zb Real execution & At least one \texttt{run\_python} call that exited 0 and produced real output. \\
    Perception & A case that carries a \texttt{look\_at\_*} budget cannot finalise a positive verdict without having looked at the raw evidence at least once. \\
    \zb Key numbers & The decisive numbers the code printed, reported as a structured record. \\
    Provenance & Every reported number must appear in the text of a real \texttt{run\_python} output. \\
    \zb Real data & Some run must have loaded the actual data rather than hand-coded rows. \\
    Reproducibility & At least $60\%$ of the reported numbers present in the union of the run's real outputs. \\
    \zb Multiple tests & Two or more reported $p$-values require a stated count of every test run and the correction applied to that count. \\
    Circularity & Whether the predictor derives from the same representation whose behaviour it predicts, and how that is handled. \\
    \zb Full battery & At least 4 analyses, each with a saved figure: primary, baseline, ablation, mechanism, breakdown, sensitivity. \\
    Lead selection & The headline must name an existing analysis, be listed first, and not also appear among the demoted ones. \\
    \zb Lead significance & A lead whose $p$-values are all $\geq 0.05$ is rejected unless the verdict is itself \emph{null} or \emph{insufficient}. \\
    Demotion & A non-significant analysis that is not the lead must be demoted, which keeps it in the trace and out of the manuscript. \\
    \zb Correction base & The stated correction count must cover the demoted analyses too, so demoting cannot shrink the denominator. \\
    \grpspan{2}{Claim check (writeup)}
    Traceability & Every number in the drafted text is matched against the grounded set derived from the experiment record. \\
    \zb Guarded revision & The prose-polish pass is reverted wholesale if it alters a number, a citation, a claim, or a model name. \\
    Forbidden claims & The over-reaching statements the thesis planner listed for this paper are removed at the output layer, deterministically. \\
    \zb Compilation & The manuscript must compile. Four fallbacks follow in order: a repair pass, the pre-polish draft, a citation-free build, and the unskinned base template. \\
    \bottomrule
  \end{tabularx}
\end{table}

\begin{table}[!t]
  \centering
  \setlength{\tabcolsep}{5pt}\renewcommand{\arraystretch}{1.06}
  \caption{Check rejections across the 36 primary runs. The two count columns differ because a single run can be rejected multiple times by the same condition. The checks refused 115 finalize attempts in total, and only 4 of the 36 runs reached both stage exits without ever being sent back. The dominant single cause is result selection: in 26 runs, the agent tried to present a non-significant analysis as a finding and was made to demote it.}
  \label{tab:gate-fired}
  \begin{tabularx}{\linewidth}{@{}>{\raggedright\arraybackslash}p{2.8cm} X r r@{}}
    \toprule
    \textbf{Condition} & \textbf{What fired} & \textbf{Rejections} & \textbf{Runs} \\
    \midrule
    \grpspanA{4}{Idea check (ideation): 28 rejections in 23 of the 36 runs}
    Schema & a required ideation field left empty & 12 & 10 \\
    \zb Effective sample & the decisive-event estimate left empty & 4 & 4 \\
    Novelty evidence & search evidence for the selected idea left empty & 3 & 3 \\
    \zb Visual audit & images inspected but the audit left empty & 3 & 3 \\
    Claim scope & what the data cannot establish left empty & 3 & 3 \\
    \zb Novelty language & absolute-novelty phrasing in the proposal & 2 & 2 \\
    Breadth & fewer than 5 screened candidates & 1 & 1 \\
    \grpspan{4}{Rigour check (experiment): 87 rejections in 32 of the 36 runs}
    Demotion & a non-significant analysis left in the paper & 51 & 26 \\
    \zb Schema & verdict left empty & 21 & 16 \\
    Lead selection & lead unset, not listed first, or a bad demotion name & 9 & 8 \\
    \zb Schema & no key numbers reported & 3 & 3 \\
    Provenance & a reported number absent from real standard output & 1 & 1 \\
    \zb Perception & finalised without looking at the raw evidence & 1 & 1 \\
    Multiple tests & the stated test count missing or under-counted & 1 & 1 \\
    \midrule
    \textbf{Total} & & \textbf{115} & \\
    \bottomrule
  \end{tabularx}
\end{table}

The distribution in \cref{tab:gate-fired} indicates where an autonomous researcher most often goes wrong.
Two thirds of all rejections concern result selection rather than arithmetic: the agent had run a broad battery, one analysis came out non-significant, and the draft still treated it as a finding.
Outright fabrication is rare, with a single provenance rejection across 36 runs; this is expected if reported numbers are normally copied from real standard output.
Most of the remainder involves schema discipline, where a field the manuscript later depends on was simply left blank.

\section{Run-level execution statistics}
\label{app:runstats}
\Cref{tab:runstats} reports the composition of a single system run, measured over the 36 primary runs by replaying their stored traces.
Ideation and experiment differ sharply in character.
Ideation is search-heavy and perception-heavy, spending most of its calls on literature searches and raw item inspection, and it never runs code.
Experiment is execution-heavy, averaging $31.8$ \texttt{run\_python} calls per run, and it returns to the raw evidence only occasionally because the observation that shaped the question has already been made.
\Cref{tab:outcomes} reports what those runs concluded.

\begin{table}[!t]
  \centering
  \setlength{\tabcolsep}{5pt}\renewcommand{\arraystretch}{1.06}
  \caption{Run composition across the 36 primary runs. Counts represent issued tool calls recovered from each run's stored trace. The step budget is 24 for ideation and 50 for experiment, and no run exhausted either.}
  \label{tab:runstats}
  \begin{tabularx}{\linewidth}{@{}X *{3}{c} !{\hspace{4pt}\color{nusgray!55}\vrule\hspace{4pt}} *{3}{c}@{}}
    \toprule
    & \multicolumn{3}{c}{\textbf{Ideation}} & \multicolumn{3}{c}{\textbf{Experiment}} \\
    \cmidrule(lr){2-4}\cmidrule(lr){5-7}
    \textbf{Per run} & \textbf{Mean} & \textbf{Median} & \textbf{Max} & \textbf{Mean} & \textbf{Median} & \textbf{Max} \\
    \midrule
    Agent steps & 8.8 & 9 & 19 & 36.0 & 37 & 49 \\
    \zb Tool calls, all kinds & 19.9 & 16.5 & 49 & 37.2 & 38 & 51 \\
    Code executions (\texttt{run\_python}) & 0.0 & 0 & 0 & 31.8 & 33 & 47 \\
    \zb Literature search calls & 8.7 & 9 & 10 & 0.0 & 0 & 0 \\
    Perception calls, all channels & 8.4 & 4.5 & 38 & 1.0 & 0 & 7 \\
    \zb \quad of which visual (\texttt{look\_at\_*}) & 4.5 & 3 & 18 & 0.9 & 0 & 7 \\
    Exit-check rejections & 0.8 & 1 & 4 & 2.4 & 2 & 7 \\
    \bottomrule
  \end{tabularx}
\end{table}

\begin{table}[!t]
  \centering
  \setlength{\tabcolsep}{6pt}\renewcommand{\arraystretch}{1.06}
  \caption{Outcomes of the 36 primary runs. The verdicts are the system's own, taken from the verified experiment record. Demoted analyses were executed and remain in the trace, but the claim check excludes them from the manuscript.}
  \label{tab:outcomes}
  \begin{tabularx}{\linewidth}{@{}X r r@{}}
    \toprule
    \textbf{Outcome over the 36-case suite} & \textbf{Count} & \textbf{Share} \\
    \midrule
    \grpspanA{3}{Self-reported verdict}
    Supported, the pre-specified hypothesis held & 16 & 44\% \\
    \zb Mixed, part of the hypothesis held & 17 & 47\% \\
    Refuted, the pre-specified hypothesis did not hold & 2 & 6\% \\
    \zb No verdict, the experiment stage exhausted its step budget & 1 & 3\% \\
    \grpspan{3}{Artifacts produced}
    Manuscripts drafted in full, with their figures & 36 & 100\% \\
    \zb Experiment stages that exited through the rigour check & 35 & 97\% \\
    Manuscripts scored by the full 2-judge panel & 36 & 100\% \\
    \grpspan{3}{Analyses per run}
    Analyses carried into the manuscript & 265 & 7.4 / run \\
    \zb Analyses demoted to the trace & 67 & 1.9 / run \\
    \bottomrule
  \end{tabularx}
\end{table}

Two of these numbers deserve comment.
The 2 refuted and 17 mixed verdicts are not system failures but the intended behaviour of the rigour check, which admits an honest negative as a valid exit and forbids promoting a weak result to the headline.
The 67 demoted analyses reflect this same mechanism, since roughly a fifth of everything the system computed was run, found wanting, and deliberately kept out of the paper.

\section{The perception layer}
\label{app:tools}
\Cref{tab:tools} lists the perception tools this suite exposed and how heavily each was used across the 36 primary runs.
The table reveals two properties of the design.
First, a modality is never forced into an image.
Signals, audio, video, 3-D structures, and trajectories each expose a native reader that returns numbers in the modality's own terms alongside a visual reader that renders the artifact for inspection, allowing the agent to choose between them.
Second, the visual channel is used substantively rather than incidentally, since $173$ of the $337$ perception calls went to a \texttt{look\_at\_*} tool and every tool listed was invoked at least once.

\begin{table}[!t]
  \centering
  \setlength{\tabcolsep}{5pt}\renewcommand{\arraystretch}{1.06}
  \caption{The perception layer. A modality automatically unlocks its tools from the specification file, requiring no per-discipline registration. \emph{Cases} indicates how many of the 36 primary runs had the tool available, and \emph{Calls} indicates how many times it was invoked.}
  \label{tab:tools}
  \begin{tabularx}{\linewidth}{@{}l l X r r@{}}
    \toprule
    \textbf{Tool} & \textbf{Modality} & \textbf{What it returns} & \textbf{Cases} & \textbf{Calls} \\
    \midrule
    \grpspanA{5}{Visual channel: render the artifact, then look at it}
    \texttt{look\_at\_image} & image & the VLM's reading of specific image files & 11 & 49 \\
    \zb \texttt{look\_at\_signal} & signal & one time-series panel per channel: onsets, bursts, envelopes & 6 & 47 \\
    \texttt{look\_at\_3d} & 3-D & rendered XY, XZ and YZ projections of a cloud or mesh & 5 & 40 \\
    \zb \texttt{look\_at\_table} & table & shape, columns, dtypes, head and summary statistics & 1 & 22 \\
    \texttt{look\_at\_audio} & audio & the rendered waveform and spectrogram & 4 & 17 \\
    \zb \texttt{look\_at\_video} & video & a sample of frames, inspected together & 3 & 13 \\
    \texttt{look\_at\_trajectory} & trajectory & the path coloured by time, plus its speed profile & 3 & 7 \\
    \grpspan{5}{Native channel: read the modality in its own terms, no image}
    \texttt{analyze\_signal} & signal & trend, dominant FFT frequencies, peaks, statistics & 6 & 40 \\
    \zb \texttt{analyze\_audio} & audio & duration, rate, RMS, spectral centroid, zero-crossing & 4 & 39 \\
    \texttt{analyze\_3d} & 3-D & point count, bounding box, centroid, extent, PCA axes & 5 & 21 \\
    \zb \texttt{read\_trace} & trace & the ordered sequence of a track or an agent run log & 2 & 20 \\
    \texttt{analyze\_trajectory} & trajectory & path length, displacement, straightness, speed, turning angles & 3 & 19 \\
    \zb \texttt{analyze\_video} & video & frame count, rate, resolution, frame-difference motion & 3 & 3 \\
    \bottomrule
  \end{tabularx}
\end{table}

\section{Task specification for a new discipline}
\label{app:spec}
Adding a discipline requires one specification file and no change to the engine.
\Cref{lst:spec} shows the file for the seismology case of \cref{sec:casestudy}, abridged to 2 of its roughly 1{,}500 members.
Four fields carry the science: the role the agent assumes, the subject the data describes, the property each record measures, and the open request.
The member list contains only the file paths and any metadata the dataset ships with.
The file names no method, hypothesis, or analysis, and the engine reads no part of it as a special case.
The \texttt{modality} field, or the file extension when it is absent, unlocks the perception tools of \cref{tab:tools}.

\prmtitle{Task specification: the seismology case}
\begin{lstlisting}[style=prompt,caption={The complete task specification for the seismology case, abridged to 2 of its roughly 1,500 members. The engine reads nothing else about the discipline.},label={lst:spec}]
{
  "role": "a seismologist analyzing three-component broadband seismic waveforms",
  "subject": "a 60-second three-component (E, N, Z) seismogram sampled at 100 Hz (a 3 x 6000
              array) recorded at a broadband seismic station (the STEAD catalogue)",
  "property": "ground-motion amplitude over time on three orthogonal components; each trace is
               catalogued as either a local earthquake or ambient noise; earthquake traces
               additionally carry source magnitude, epicentral distance (km) and depth (km), and
               every trace carries station network, station code and instrument channel metadata",
  "request": "Find a concrete, novel, testable question this data can answer, decide your own
              method, run real code on the waveforms, and produce a short publishable paper.",
  "members": [
    {"idx": 0, "file": "data/seis_0000.npy", "label": "earthquake", $"modality": "signal"$,
     "network": "NN", "station": "COLR", "channel": "HH",
     "magnitude": 1.2, "distance_km": 17.2, "depth_km": 8.6},
    {"idx": 1, "file": "data/seis_0001.npy", "label": "noise", $"modality": "signal"$,
     "network": "AG", "station": "LCAR", "channel": "HH",
     "magnitude": null, "distance_km": null, "depth_km": null}
  ]
}
\end{lstlisting}

\section{Field-specific writeup specifications}
\label{app:writeup}
The writeup stage supports 5 structural specifications, resolved from the case specification or inferred from the subject.
\Cref{tab:venue-specs} shows their skeletons.
Each section additionally receives a word budget, a paragraph count, and a paragraph-level outline.
The drafting pass expands this outline one section at a time, using only the slice of the experiment record allocated to that section.
The structural differences between these skeletons reflect actual field conventions: a machine-learning paper includes Related Work and Limitations, a biomedical paper puts Methods last, and a chemistry paper merges Results with Discussion.

\begin{table}[!t]
  \centering
  \setlength{\tabcolsep}{5pt}\renewcommand{\arraystretch}{1.12}
  \caption{The 5 structural specifications that the writeup stage can resolve to. The abstract column shows the word range for a single-paragraph abstract.}
  \label{tab:venue-specs}
  \begin{tabularx}{\linewidth}{@{}>{\raggedright\arraybackslash}p{2.7cm} X c@{}}
    \toprule
    \textbf{Style} & \textbf{Section order} & \textbf{Abstract} \\
    \midrule
    Machine learning & Introduction, Related Work, Method, Experiments, Conclusion, Limitations & 150--220 \\
    \zb Biomedical & Introduction, Results, Discussion, Methods & 150--200 \\
    Earth \& space & Introduction, Data, Methods, Results, Discussion, Conclusions & 150--250 \\
    \zb Physics & Introduction, Theory and Methods, Results, Discussion, Conclusion & 150--250 \\
    Chemistry & Introduction, Experimental Section, Results and Discussion, Conclusions & 150--250 \\
    \bottomrule
  \end{tabularx}
\end{table}

\section{Stage prompts}
\label{app:prompts}
Ideation and experiment are each driven by a single system prompt assembled at run time from the case specification, allowing one template to serve every discipline.
The role and request come from \cref{lst:spec}, and the list of available perception tools comes from the detected modalities.
\Cref{lst:p1,lst:p2} reproduce the process and grounding sections of these two prompts verbatim, omitting the tool inventory and role interpolation.
The writeup stage has no comparable single text because it is prompted one section at a time using the structural specification in \cref{app:writeup} and the slice of the experiment record available to that section.

Two properties of these prompts are worth noting alongside the checks of \cref{app:checks}.
First, every demand the prompt makes is also a predicate.
Because the prompt states what is wanted and the exit check refuses the stage when it is missing, the instruction does not depend on the model choosing to comply.
Second, the prompts never name a discipline, a modality, or a method.
They describe how to conduct research, and the specification file supplies what the research is about.

\prmtitle{Ideation stage: system prompt}
\begin{lstlisting}[style=prompt,caption={Ideation system prompt, process and grounding sections, verbatim. The tool list and the role string are interpolated from the case specification.},label={lst:p1}]
@PROCESS@ (keeps the idea non-trivial, novel, AND actually runnable -- do not skip it):
1. INSPECT THE MATERIALS: list_materials (and use the perception tools on a few representative
   items) to identify WHAT KIND of data this is and what is actually in it.
2. KNOWN LANDSCAPE: a SMALL number of FOCUSED literature searches (about 3-6) to establish what
   is already well-established, so you deliberately AVOID it.
3. FIND THE QUESTION: from what the materials actually contain, decide the most concrete, novel,
   testable question this data can genuinely support. Do NOT force a question the data cannot
   sustain.
4. BRAINSTORM >=5 candidate research projects. Use what you OBSERVED in the materials to
   GENERATE candidates from concrete patterns, not literature alone. Self-screen EACH on
   novelty_risk (already published / obvious? low / med / high).
5. FEASIBILITY (be brutally honest -- this is a $HARD GATE$): rate each candidate
   fully_computational -- can it be carried out END-TO-END on a computer with NO wet-lab
   experiment? Also note falsifiable and statistical_power at this sample size.
6. SELECT the ONE candidate that is GENUINELY NOVEL and fully_computational=YES -- the goal is
   NOVEL AND feasible, NOT the safest option; also falsifiable, and 'valuable even if it fails',
   preferring one whose core hypothesis was inspired by inspecting the materials rather than
   literature alone.
7. Develop the selected candidate into a full, falsifiable proposal and call finalize_idea.

@GROUNDING RULES@ (a proposal that violates these is NOT ready -- the exit gate enforces them):
- VISION: when you look at an item you are ALSO shown its GIVEN label; RECONCILE your read with
  it. If they disagree, do NOT proceed on your read -- flag it, re-check the id / label, lower
  confidence. Never name a class outside the given label space, and never characterize a group
  you did not view.
- NOVELTY: run >=3 focused searches incl. one aimed at your SELECTED idea specifically; phrase
  novelty as 'based on this search, appears under-explored' -- $NEVER$ 'unstudied / first / no
  prior work'.
- CLAIM SCOPE: separate what the data DIRECTLY shows from a broader mechanism or causal claim it
  cannot prove; keep the research_question and the minimal_publishable_claim on the supported side.
- STAT POWER: estimate from the REAL data counts how many DECISIVE events your key test needs;
  if they will be rare, use CONTINUOUS signals or the full dataset -- do not rely on a handful of
  hard events.
- NO LEAKAGE: no step may use ground-truth labels at inference / decision time; any
  label-conditioned selection is oracle / upper-bound only and needs a label-agnostic counterpart
  plus its false-positive rate.
\end{lstlisting}

\prmtitle{Experiment stage: system prompt}
\begin{lstlisting}[style=prompt,caption={Experiment system prompt: the battery, the anti-fabrication rule and the anti-HARKing rule, verbatim.},label={lst:p2}]
@A FULL STUDY@ = this battery (do every one the data can support; aim for >=5 analyses and ~8
figures / tables):
  (1) PRIMARY: the pre-specified test of your hypothesis. It MAY come out weak or fail -- that
      is fine and normal; do NOT force it to look like a win.
  (2) BASELINE: contrast vs a simpler / alternative method or a trivial baseline.
  (3) ABLATION: remove or vary a component to show which part drives the result.
  (4) MECHANISM: dig into WHY the effect arises, not just that it does.
  (5) BREAKDOWN: split by group / class / condition / time -- where it holds and where not.
  (6) SENSITIVITY: vary the key thresholds / hyperparameters / subsample and show the finding
      is stable.
  (7) LEAKAGE / GROUPING: if the data carries ANY grouping id (source / event / subject /
      patient / site / recording / network), you MUST evaluate under a group-disjoint or
      leave-one-group-out split, and report the group-out metric next to the random-split one.

@IRON RULE@ -- $NO FABRICATION$: every number you report MUST come from real run_python output.
NEVER hard-code, invent, or 'expand' data rows -- LOAD the real data. If the data genuinely
cannot support the test, run what you can and report verdict='infeasible' with what you found.

@RIGOR@ (the exit gate enforces these):
- MULTIPLE COMPARISONS: if you run several statistical tests, COUNT every test you actually ran
  and correct for THAT count (Bonferroni / FDR); report which results survive. Do not under-count
  the tests.
- INDEPENDENCE: if your predictor and your outcome come from the SAME model or representation,
  the correlation may be circular -- use an INDEPENDENT predictor or explicitly scope the claim
  to 'within this representation'.
- HONEST POWER: a sub-result resting on very few events is 'suggestive', NOT a confirmed
  finding -- say so.

@REFRAME@ (do this at the END, before you finalize): pick the LEAD = the single STRONGEST
SUPPORTED analysis in your battery. $HARKing guard$: a LEAD is only allowed if it (a) survives
your multiple-comparison correction, (b) has a clear effect size, and (c) is not just the best of
many noisy tries; if nothing clears this bar, report verdict='mixed' or 'null' honestly -- do NOT
promote a weak result. Put any failed or weak analysis (including a failed PRIMARY) in 'demoted';
the paper does not mention it, it only lives in the trace.
\end{lstlisting}

\end{document}